\documentclass{nature}
\usepackage[utf8]{inputenc} 
\usepackage[T1]{fontenc}    
\usepackage{hyperref}       
\usepackage{url}            
\usepackage{booktabs}       
\usepackage{amsfonts}       
\usepackage{amsmath}       
\usepackage{nicefrac}       
\usepackage{microtype}      
\usepackage{xcolor}         
\usepackage{graphicx}
\usepackage{bm}
\usepackage{ulem}
\usepackage{utfsym}
\usepackage{lineno}

\usepackage[font=small]{caption}

\title{Genes in Intelligent Agents}

\author{Fu Feng$^{1,2}$, Jing Wang$^{1,2}$, Xu Yang$^{1,2}$ \& Xin Geng$^{1,2}$\footnote{Corresponding author. Email: xgeng@seu.edu.cn}}

\begin{document}
\captionsetup[figure]{name={Fig.}, labelsep=space}
\maketitle

\begin{affiliations}
 \item School of Computer Science and Engineering, Southeast University, Nanjing, China
 \item Key Laboratory of New Generation Artificial Intelligence Technology and Its Interdisciplinary Applications (Southeast University), Ministry of Education, China
\end{affiliations}

\begin{abstract}
The genes in nature give the lives on earth the current biological intelligence through transmission and accumulation over billions of years. Inspired by the biological intelligence, artificial intelligence (AI) has devoted to building the machine intelligence. Although it has achieved thriving successes, the machine intelligence still lags far behind the biological intelligence. 
The reason may lie in that animals are born with some intelligence encoded in their genes, but machines lack such intelligence and learn from scratch.
Inspired by the genes of animals, we define the ``genes'' of machines named as the ``learngenes'' and propose the Genetic Reinforcement Learning (GRL). GRL is a computational framework that simulates the evolution of organisms in reinforcement learning (RL) and leverages the learngenes to learn and evolve the intelligence agents. 
Leveraging GRL, we first show that the learngenes take the form of the fragments of the agents' neural networks and can be inherited across generations. 
Second, we validate that the learngenes can transfer ancestral experience to the agents and bring them instincts and strong learning abilities. 
Third, we justify the Lamarckian inheritance of the intelligent agents and the continuous evolution of the learngenes. 
Overall, the learngenes have taken the machine intelligence one more step toward the biological intelligence.
\end{abstract}

The evolution over 3.5 billion years has given the lives on earth the current biological intelligence\cite{braga2017emperor, oro2004evolution}, such as the social learning of fruit flies\cite{danchin2018cultural} and the numerical cognition in honeybees\cite{howard2019numerical}. 
Inspired by the biological intelligence, the field of artificial intelligence (AI) is devoted to building the machine intelligence\cite{hassabis2017neuroscience}. The artificial neural networks (ANN) simulate our brains from the perspective of information processing\cite{matsuo2022deep, kriegeskorte2015deep}. Reinforcement learning (RL)\cite{sutton2018reinforcement} mimics the learning process of animals, where intelligent agents make decisions by interacting with the environments in a trial-and-error approach\cite{gronauer2022multi, nguyen2020deep}.

AI has already achieved thriving successes. For example, AlphaGo has beaten the top-ranked Go players\cite{silver2016mastering, silver2017mastering}. ChatGPT, a representative large language model (LLM), can generate sophisticated and structured contents\cite{van2023chatgpt}.
Indeed, the number of neurons in the mainstream LLMs (e.g., 175 billion parameters in GPT-3)\cite{abio2023ai} has already matched that of human brains (86 billion neurons)\cite{landhuis2017neuroscience}.
However, there still exists a significant gap between the machine intelligence and biological intelligence. Spiders are born with the ability to spin webs\cite{krink1997analysing}, and newborn colts can walk in a short period\cite{gorissen2017development}. These are innate behaviors of animals from evolution\cite{seung2012connectome}, which enable animals to fast adapt to their environments\cite{wong2015behavioral, sih2011evolution}. In comparison, machines lack such instincts and generally learn from scratch\cite{tan2022rlx2, riedmiller2018learning, bakhtin2021no}, which is extremely time-consuming and inefficient\cite{agarwal2022reincarnating, li2018a2, wang2020reinforcement}.

We believe the gap is primarily due to the absence of an inherent mechanism in machines, like the genes in organisms. Gene plays a crucial role in the biological evolution. It can transmit traits between individuals through inheritance \cite{bohacek2015molecular, waddington1942canalization} and accumulate dominant mutations of populations through evolution\cite{andreev2022oldest, zador2023catalyzing}. 
That means our brains come equipped with the prior knowledge accumulated over billions of years of evolution\cite{oro2004evolution, lillicrap2020backpropagation}. Hence, the nervous system of a newborn is not randomly initialized but wired up under a blueprint encoded in the genes, such as the connected neurons and the strengths of these connections\cite{zador2019critique, robinson2017epigenetics}. 
In this way, animals have a prior understanding of their world with survival mechanisms encoded in their genes\cite{moravec1988mind}, so they are born to have some instincts. In contrast, machines, such as ANN and intelligent agents in RL, are initialized without any knowledge from the genes as animals. 

\begin{figure}[tb]
  \centering
  \includegraphics[width=39em]{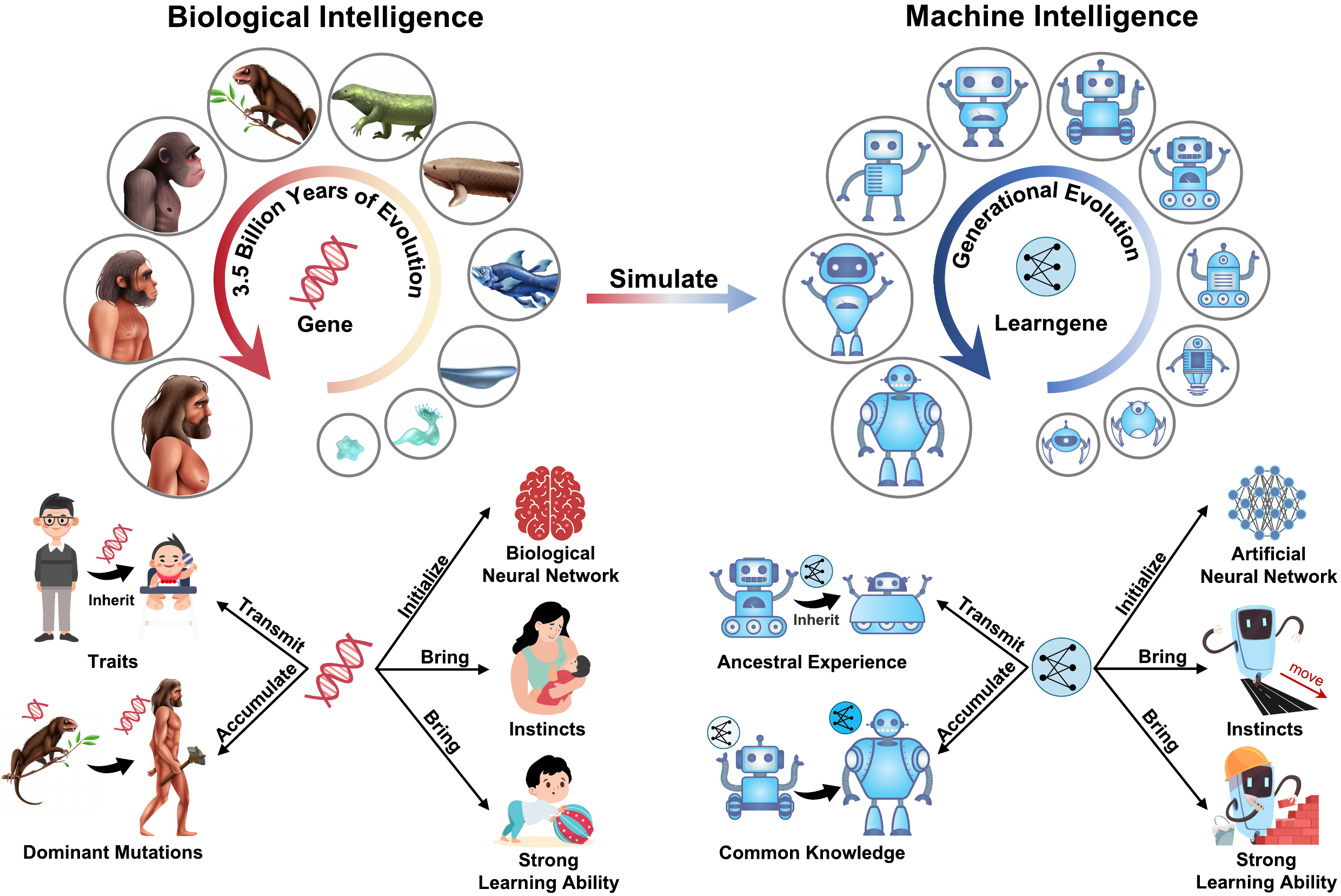}
  \caption{\textbf{Biological intelligence and machine intelligence.} The genes in nature have evolved over 3.5 billion years before giving lives on earth today's biological intelligence. Genes can transmit traits, accumulate dominant mutations, encode how our brain are wired up, and bring us instincts and strong learning abilities. To simulate such an evolution process in machines and achieve machine intelligence, we define the ``genes'' for machines called the ``learngenes''. The learngenes, similar to the genes of organisms, can transmit ancestral experience, accumulate common knowledge, initialize ANN (i.e., the ``brain'' of machines), and bring machines instincts and strong learning abilities.}
  \label{fig:intro}
\end{figure}

Currently, a mainstream way toward the biological intelligence in machines is the brain-inspired computing\cite{poo2016china}. It successfully copies the neuron structure of human brain into machines\cite{lv2022post,mehonic2022brain}, which is a highly optimized natural product through billions of years of evolution\cite{gilbert2005genetic, roberts2022evolution, niven2016evolving}.
However, there exists a possible challenge that the results of evolution for biological intelligence, especially the neuron structure, may not suit for the machine intelligence and necessarily bring the biological intelligence to machines\cite{brodner2018super, luo2021architectures}.
Another possible way to achieve better machine intelligence is to simulate the process of biological evolution\cite{hasson2020direct}, which allows machines to evolve on their own paces and achieve the suitable intelligent forms, such as the structures and morphologies\cite{gupta2021embodied}. 

\begin{figure}[tb]
  \centering
  \includegraphics[width=39em]{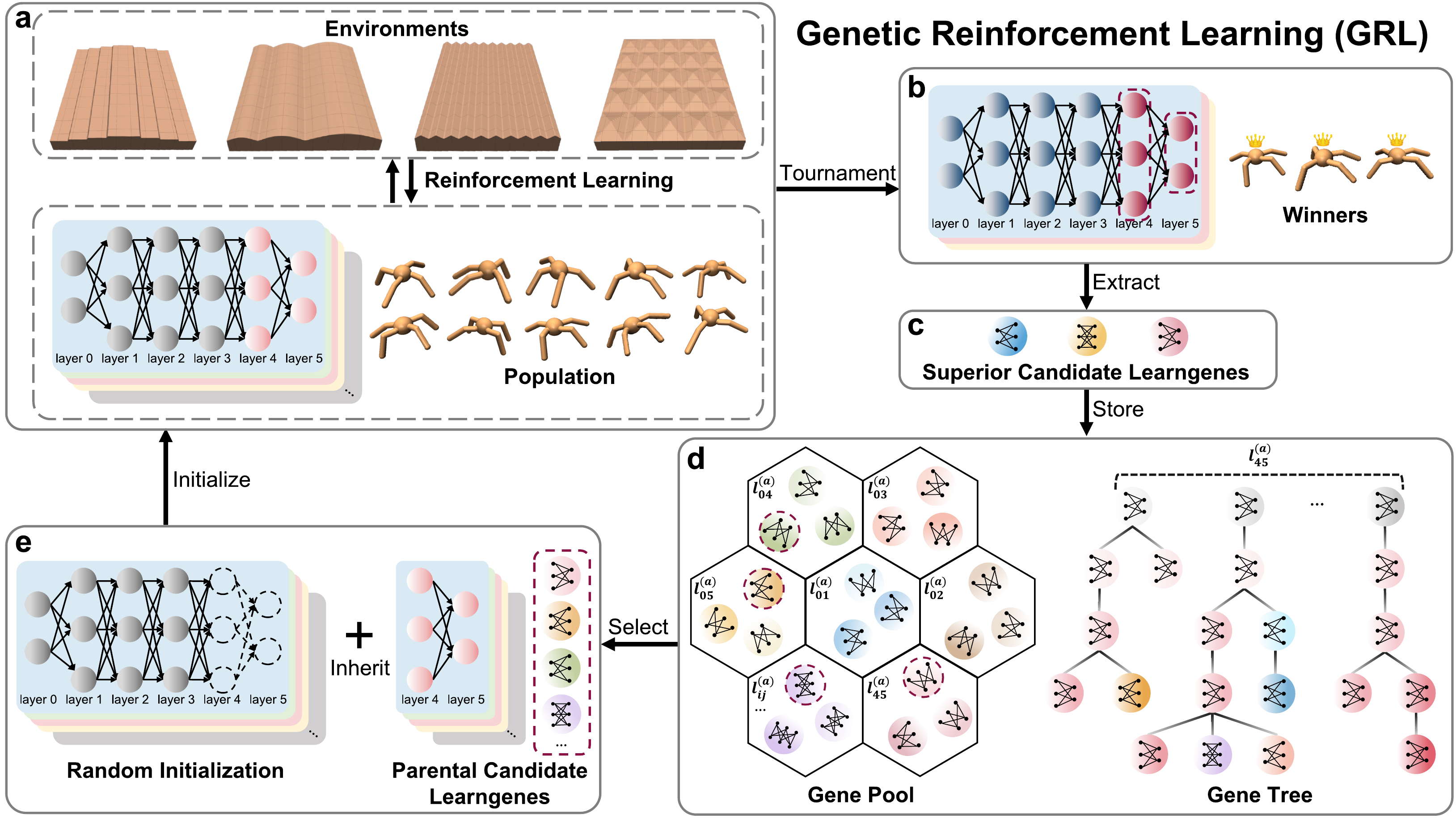}
  \caption{\textbf{GRL overview.} GRL is a computational framework to train the agents and evolve the learngenes. This framework is a loop of five key components. \textbf{a} Reinforcement learning where the agents learn by interactions with the environments. \textbf{b} Tournaments for selecting the winners in the population. 
  \textbf{c} Superior candidate learngenes extracted from the winners. \textbf{d} Gene Pool and Gene Tree that store the superior learngenes and record the kinship of the learngenes, respectively. \textbf{e} Initialization of the population of the next generation with the learngenes in the Gene Pool.}
  \label{fig:GRL}
\end{figure}

\begin{figure}[tb]
  \centering
  \includegraphics[width=39em]{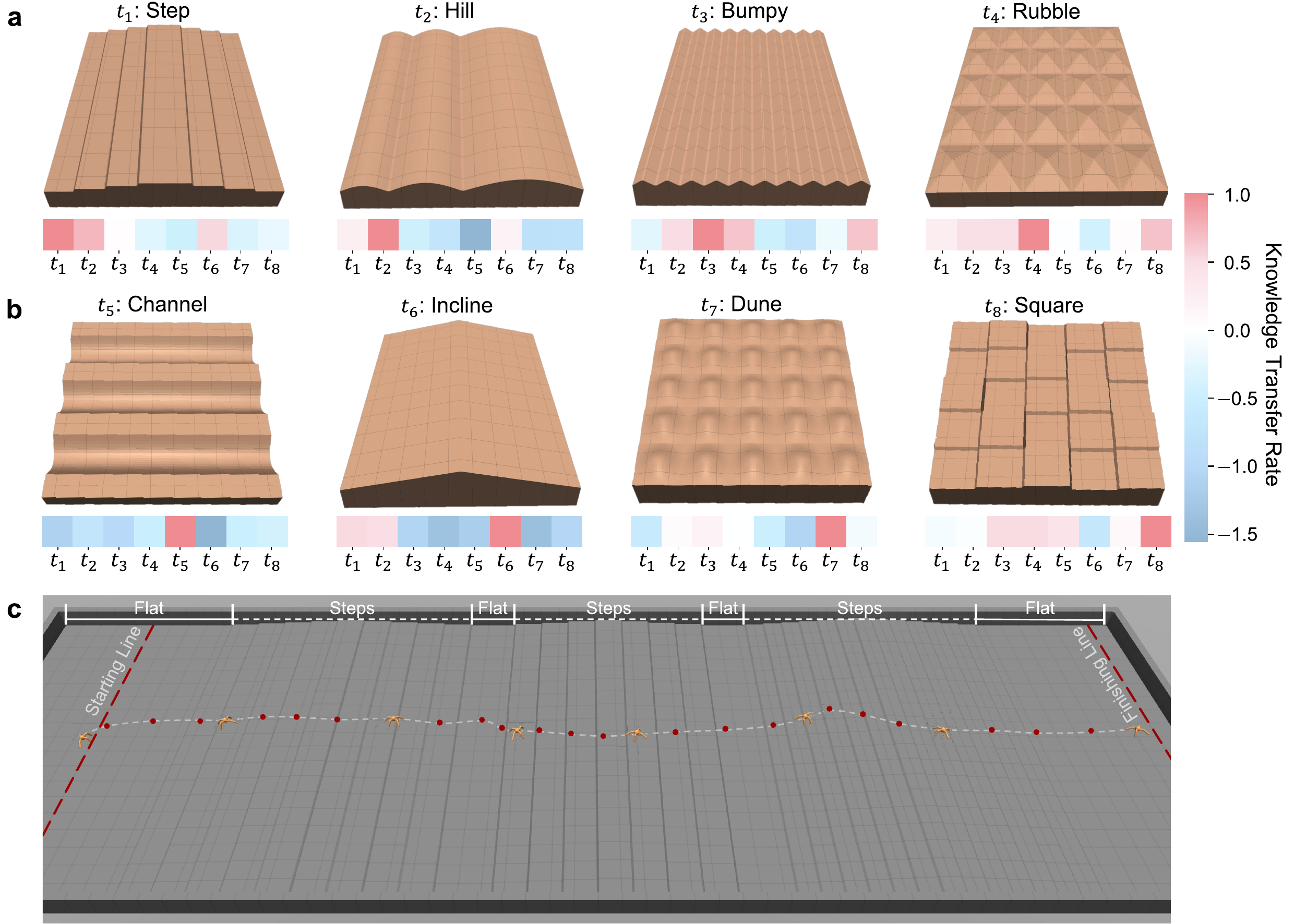}
  \caption{\textbf{Obstacles in variable terrain tasks.} We use four training obstacles (\textbf{a}) to train the agents and evolve the learngenes in GRL. All the obstacles (i.e., four training obstacles (\textbf{a}) and four new obstacles (\textbf{b})) are used to evaluate the advantages of the learngenes. The heat map under the obstacle $t_i$ denotes the knowledge transfer rate from the obstacle $t_j$ to $t_i$, which is calculated by Eq. \eqref{equ:knowledge_tranfer}. The positive value from $t_j$ to $t_i$ indicates that the knowledge acquired from $t_j$ helps the learning of $t_i$, and the negative value indicates the counterproductive knowledge between $t_j$ and $t_i$. 
  \textbf{c} An overview of the task environments with the Steps (i.e., $t_1$). The agents start from the starting line, cross the obstacles, and reach the finishing line.}
  \label{fig:8 obstacle}
\end{figure}

To facilitate the evolution process of machines, we define in this paper the genes for machines, called the ``learngenes'' (Fig. \ref{fig:intro}). Especially, the learngenes carry some ``genetic materials'' (i.e., common knowledge) for the learning of machines and work in the evolution process of machines. Since neural networks can be viewed as the ``brain'' of machines, we represent the learngenes as the fragments of the neural networks. To implement the learning and evolution of machines, we propose Genetic Reinforcement Learning (GRL) (Fig. \ref{fig:GRL}) ---a novel computational framework that leverages the learngenes to learn and evolve intelligent agents across generations. GRL enables the agents to inherit the learngenes from their ancestors, learn with better instincts, and pass on learngenes to the next generation, which simulates the evolution process of animals. 
Concretely, we places the agents in variable terrain tasks\cite{gupta2021embodied,hoffmann2011climate} (Fig. \ref{fig:8 obstacle}c) to obtain the learngenes concerning locomotion and obstacles (Fig. \ref{fig:8 obstacle}a), since locomotion profoundly influences the physiology of animals\cite{dickinson2000animals} with various obstacles simulating different living environments.

GRL makes a breakthrough at conceptualizing the ``genes'' in the intelligent agents (i.e., the learngenes) toward better intelligence and demonstrating what the learngenes bring to the agents. Our major contributions are as follows. 
First, we define the ``genes'' in the intelligent agents, which are named as the ``learngenes''. The learngenes take the form of the fragments of the neural networks, which can be inherited across the generations. 
Second, we demonstrate that the learngenes of intelligent agents have similar natures to the genes of animals. Likewise, the learngenes condense the common knowledge from ancestors, and therefore the newborn agents inheriting the learngenes show instincts, which have never been observed in machines before. 
Also, learngenes bring the agents with strong learning abilities, thus helping them quickly adapt to the new environments. 
Third, we justify that the learngenes also continually evolve in the evolution process. The evolution of the agents sufficiently satisfies the Lamarckian inheritance, where the agents consistently encode knowledge into the learngenes during the environmental interactions and pass the learngenes to the descendants, thus propelling the continual evolution of the learngenes. 
Overall, the emergence of the learngenes has taken the machine intelligence one more step toward the biological intelligence.

\section*{Results}
\label{section:sec_results}

\subsection{GRL: a computational framework for learning and evolution.} We propose Genetic Reinforcement Learning (GRL) that is a computational framework to simulate the genetic process of organisms. We perform large-scale experiments with GRL to train agents and evolve the learngenes through hundreds of generations (Fig. \ref{fig:GRL}). Specifically, the initial generation starts with a population of $n_p=50$ agents and undergoes lifetime RL in parallel. The evolution runs in a tournament-based manner after the lifetime RL of all agents, where $s=3$ agents are randomly selected (without replacement) to engage in a tournament with their average rewards as the fitness. The learngenes are formulated as the fragments (combination of layers) of the neural networks\cite{wang2022learngene}, and better candidate learngenes are extracted from the winners. In the next generalization, a new population is initialized by inheriting the candidate learngenes, and a new evolution starts. See the \nameref{section:sec_method} for details. 

\begin{figure}[tb]
  \centering
  \includegraphics[width=39em]{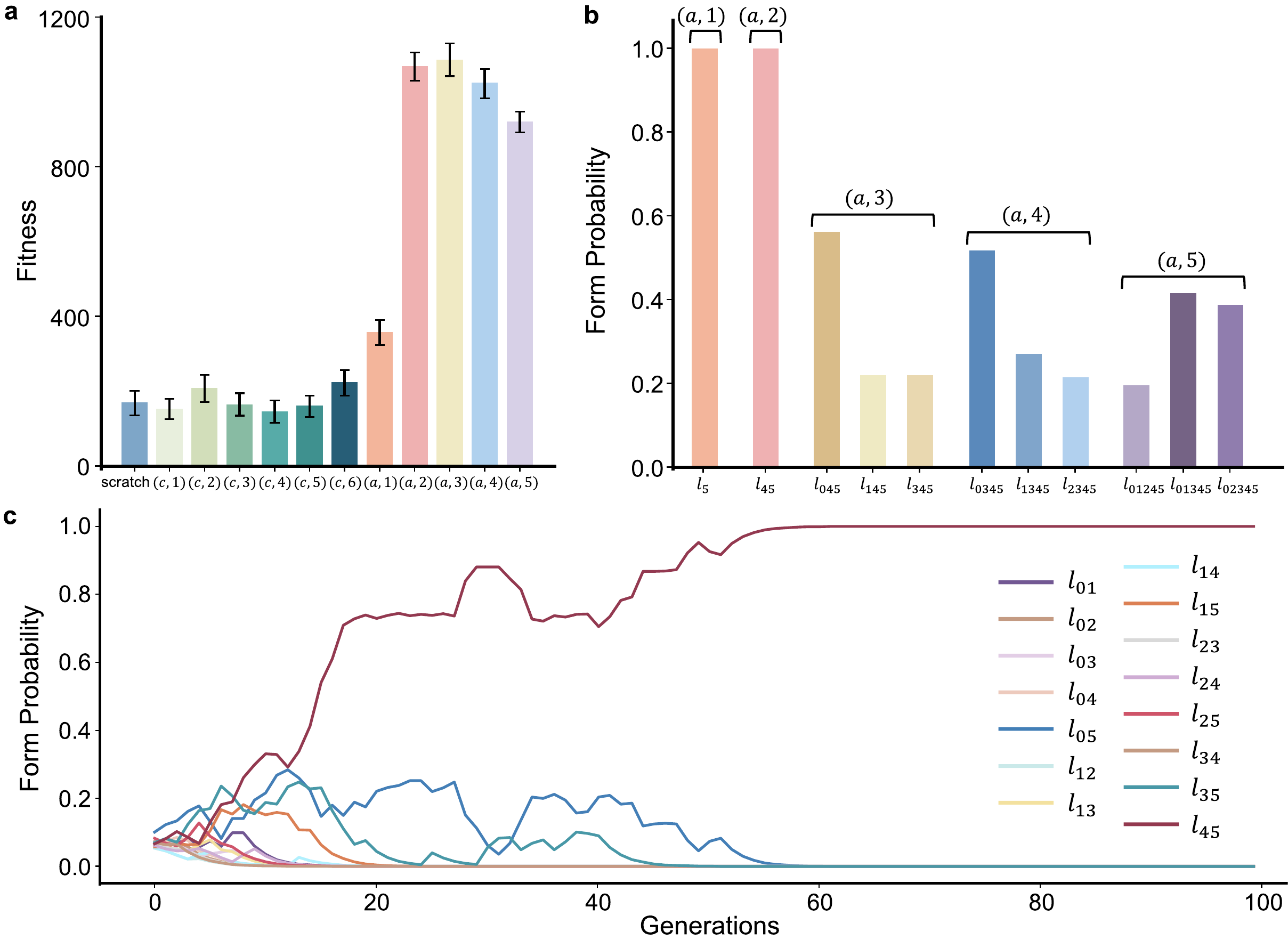}
  \caption{\textbf{Learngenes extracted by GRL.} 
  \textbf{a} The mean and 95\% bootstrapped confidence intervals of the fitness of the entire population at the end of evolution (i.e., the 100th generation) with the learngenes as the combination of $n_l$ layers from the actor network (i.e., $(a,n_l)$) or critic network (i.e., $(c,n_l)$). 
  \textbf{b} The form probability (non-zero) of each candidate learngene form in $(a, n_l)$ at the end of evolution.
  \textbf{c} Change curves of the form probability of all candidate learngene forms in $(a,2)$ (Supplementary Fig. 1 is the corresponding change curves of $(a, 1)$, $(a, 3)$, $(a, 4)$ and $(a, 5)$).}
  \label{fig:condense_gene}
\end{figure}

\subsection{Learngenes in the intelligent agents.}
\label{section:sec_gene_GEL}
The agents in GRL have the actor-critic network structure whose parameters are optimized by the Proximal Policy Optimization (PPO) algorithm\cite{schulman2017proximal}. Both the actor and critic network of an agent have six layers in our experiments.
The learngenes take the form of a combination of $n_l$ layers from the actor or critic networks ($n_l \in [1, 5]$ because we think the scale of the learngenes is less than that of the whole actor or critic networks. See section \nameref{section:sec_learngene} for details). 

The actor network is a mapping from the observations to the actions, and the critic network judges the quality of the states\cite{jiang2020structure}. Therefore, the actor network can be regarded as the ``brain'' of an agent. In addition, by transmitting the fragments of the actor network as the candidate learngenes through the inheritance, the agents gained significant benefits in terms of the fitness (Fig. \ref{fig:condense_gene}a). In comparison, the fragments of the critic network almost bring nothing to the agents' fitness (Fig. \ref{fig:condense_gene}a). As a result, we believe the learngenes exist in the actor networks of the agents. 

GRL evolves the agents with the candidate learngenes composed of $n_l$ layers from the actor network, denoted by $(a, n_l)$. All the candidate learngenes are stored in the Gene Pool. Especially, for $(a, 2)$, its corresponding candidate set for the learngene form is $\{l_{01}^{(a)}, l_{02}^{(a)}, \cdots, l_{ij}^{(a)}, \cdots, l_{45}^{(a)}\}$, where $l_{ij}^{(a)}$ is a combination of the $i$th and $j$th layers from the actor network \footnote{\label{fn:fn_omit}For simplicity, we omit the superscript $(a)$ without ambiguity and use $l_{ij}$ to represent $l_{ij}^{(a)}$.}. GRL extracts the optimal learngenes in a dynamic way according to a distribution over the candidate set, whose element (form probability) is the probability of the corresponding form as the learngene form (see section \nameref{sec:sec_condense_gene}).

Finally, only $l_{5}$ and $l_{45}$ achieve stable inheritance in the dynamic process of evolution (i.e., the form probability of $l_{5}$ and $l_{45}$ are 1 and others are 0 according to Fig. \ref{fig:condense_gene}b,c).
That is, the knowledge in $l_{5}$ and $l_{45}$ is the common knowledge among various tasks condensed from the ancestors.
As the common knowledge in $l_{45}$ results in remarkably higher fitness than that of $l_5$ (Fig.  \ref{fig:condense_gene}a), we take $l_{45}$ as the learngene form in the agents. 
Note that the knowledge in other layers may also contain some common knowledge, which is sparse and scattered (see $(a, 3)$ in Fig. \ref{fig:condense_gene}a,b). We choose the fragments of neural networks where the common knowledge concentrates as the learngene form.

For an agent $\alpha$, its parameters corresponding to $l_{ij}$ is denoted by $l_{ij}(\alpha)$\footnote{We use $l_{ij}$ to denote the candidate learngene form, and $l_{ij}(\cdot)$ to denote the candidate learngenes with the candidate learngene form $l_{ij}$.}. According to the above analysis, the candidate 
learngenes $l_{45}(\cdot)$, which are formed of $l_{45}$, are the learngenes $\mathcal{G}=l_{45}(\cdot)$ in the intelligent agents. The learngenes $\mathcal{G}$ from the Gene Pool of the $100th$ generation are the optimal learngenes $\mathcal{G}^{\star}$ that are finally extracted, that is, $\mathcal{G}^{\star} = l_{45}(\alpha^{\star})$ and $\alpha^{\star}$ is the optimal agents carrying $\mathcal{G}^{\star}$ (hereafter the learngenes inherited by the agents are all $\mathcal{G}^{\star}$). 

\subsection{Learngenes bring strong instincts to the agents.}
\begin{figure}[tb]
  \centering
  \includegraphics[width=39em]{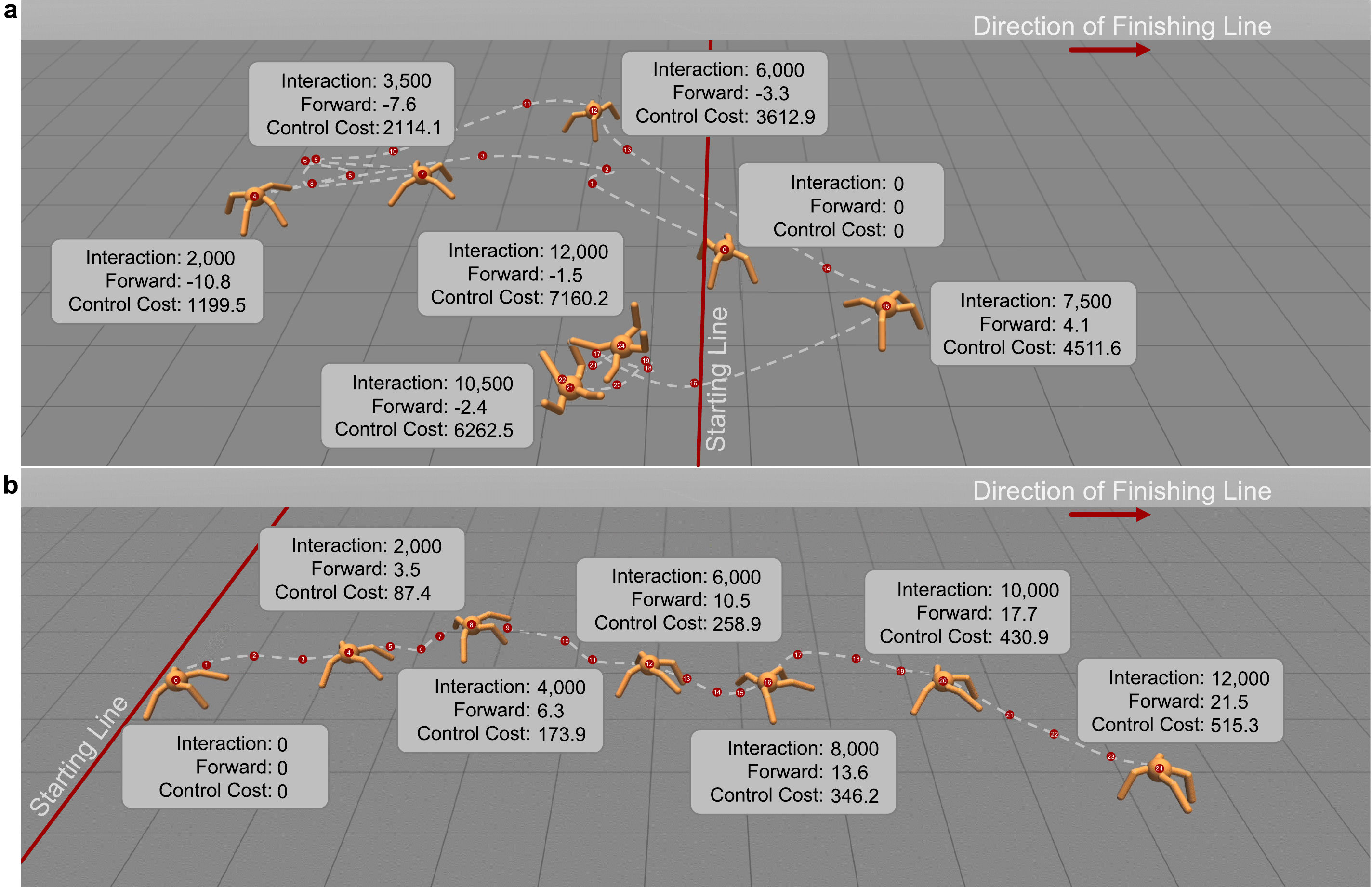}
  \caption{\textbf{Strong instincts observed in the intelligent agents.} The motion trajectory of the newborn agents that (\textbf{a}) are randomly initialized and (\textbf{b}) inherit the learngenes. 
  The locations of the agents are recorded for every 500 interactions with environments (marked by the red dots with white numbers (i.e., 0-25) above representing the order of locations). The forward distance (the larger, the better) from the starting line, as well as the control costs (the smaller, the better) for penalizing the agents taking too large actions (Eq. \eqref{equ:equ_reward}) are also recorded at some key points. Note that there was no update of any parameters during the entire interactions with environments.}
  \label{fig:instinct}
\end{figure} 

The innate behaviors of animals are encoded in their genes through evolution\cite{seung2012connectome}, which prompt them to rapidly adapt to their environments with few interactions\cite{wong2015behavioral, sih2011evolution}. For instance, spiders are born with the ability to spin webs\cite{krink1997analysing}, and a newborn colt can walk in a short period\cite{gorissen2017development}.
Surprisingly, we observed instincts in the intelligent agents trained by GRL. To the best of our knowledge, this is the first time that instinct is introduced to the machine intelligence. 

The newborn agents with random initialization have no knowledge about the environment and thus explore the environment aimlessly using large actions (Fig. \ref{fig:instinct}a). In contrast, the newborn agents inheriting the learngenes (i.e., $\mathcal{G}^{\star}$) exhibit completely different behaviors (Fig. \ref{fig:instinct}b, see Supplementary Video for details). The learngenes have condensed the knowledge on how to obtain more rewards through evolution, so the agents inheriting the learngenes will instinctively make small actions (the control costs are only 7\% of that of the randomly initialized agents) and move slowly toward the finishing line (the straighter motion trajectory and longer forward distance, Fig. \ref{fig:instinct}b).

\subsection{Learngenes help the agents quickly adapt to the new environments.}
\begin{figure}[tb]
  \centering
  \includegraphics[width=39em]{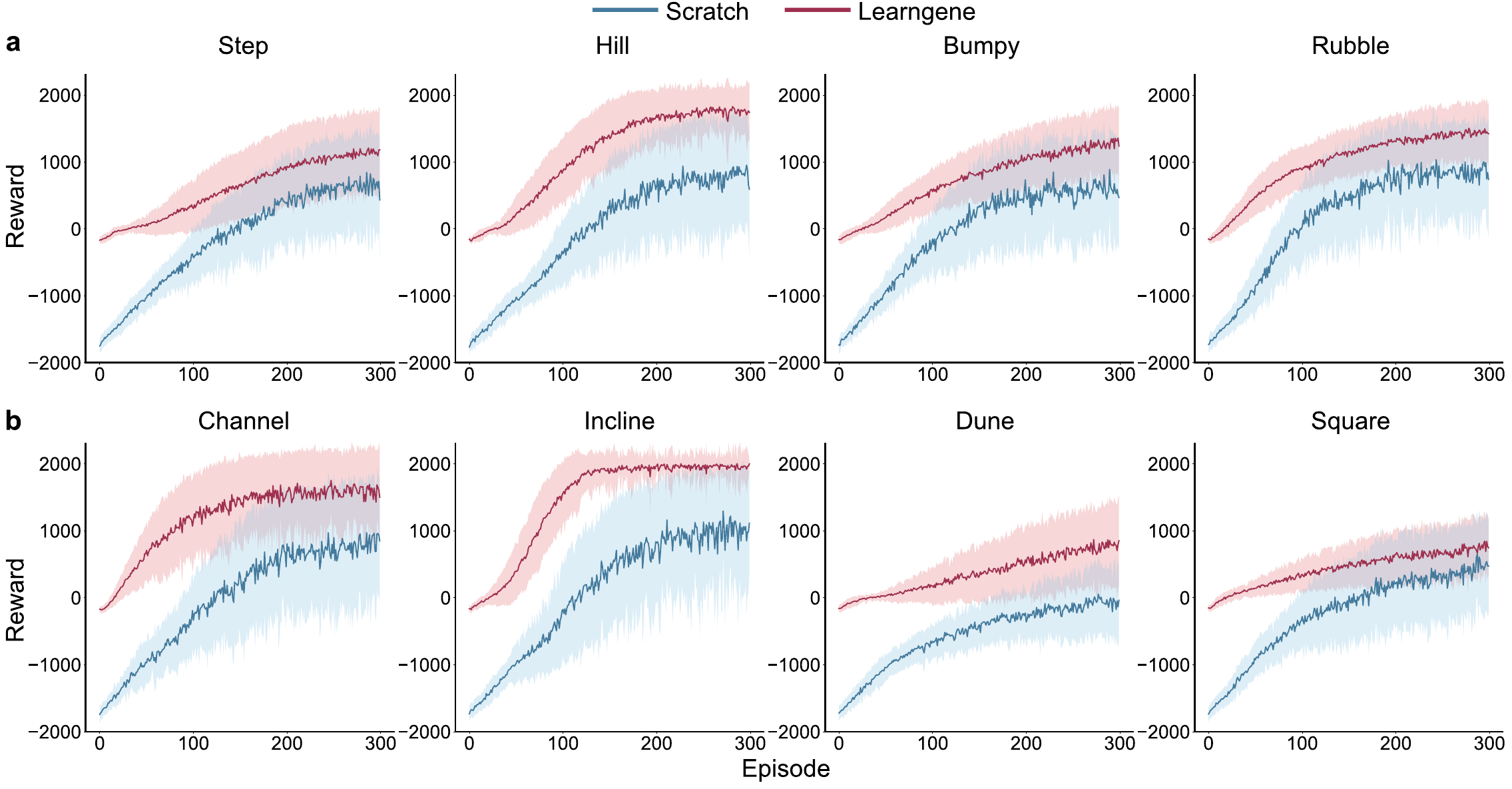}
  \caption{\textbf{Compared with learning from scratch.} The mean and standard deviation of rewards of the agents ($n=50$) trained on eight tasks (Fig. \ref{fig:8 obstacle}). The agents inheriting the learngenes all achieved higher rewards than the agents learning from scratch on \textbf{(a)} the four training obstacles used to extract the learngenes and \textbf{(b)} the four new obstacles.}
  \label{fig:scratch and gene}
\end{figure} 

In generational evolution, each agent is randomly assigned to a terrain task from four training obstacles (Fig. \ref{fig:8 obstacle}a). As a result, the learngenes extracted by GRL are not specific to one task but common for learning these four tasks. The agents inheriting the learngenes (i.e., $\mathcal{G}^{\star}$) learn faster and better than those learning from scratch (i.e., without inheriting the learngenes) on all these four training tasks since birth, and the time to reach the same rewards is shortened by at least 100 episodes (Fig. \ref{fig:scratch and gene}a, see Supplementary Video for details). 

More interestingly, even on the four new terrain tasks (Fig. \ref{fig:8 obstacle}b) that the ancestors of the agents have never seen, the learngenes also help the agents adapt fast to these terrain tasks, which remarkably surpass the agents learning from scratch (Fig. \ref{fig:scratch and gene}b, see Supplementary Video for details).
Especially on the task with the Dune obstacles, the rewards of the agents inheriting the learngenes at birth even exceed those of the agents learning from scratch during their entire lifetime.

\subsection{Learngenes transfer common knowledge between the agents.}
\begin{figure}[tb]
  \centering
  \includegraphics[width=39em]{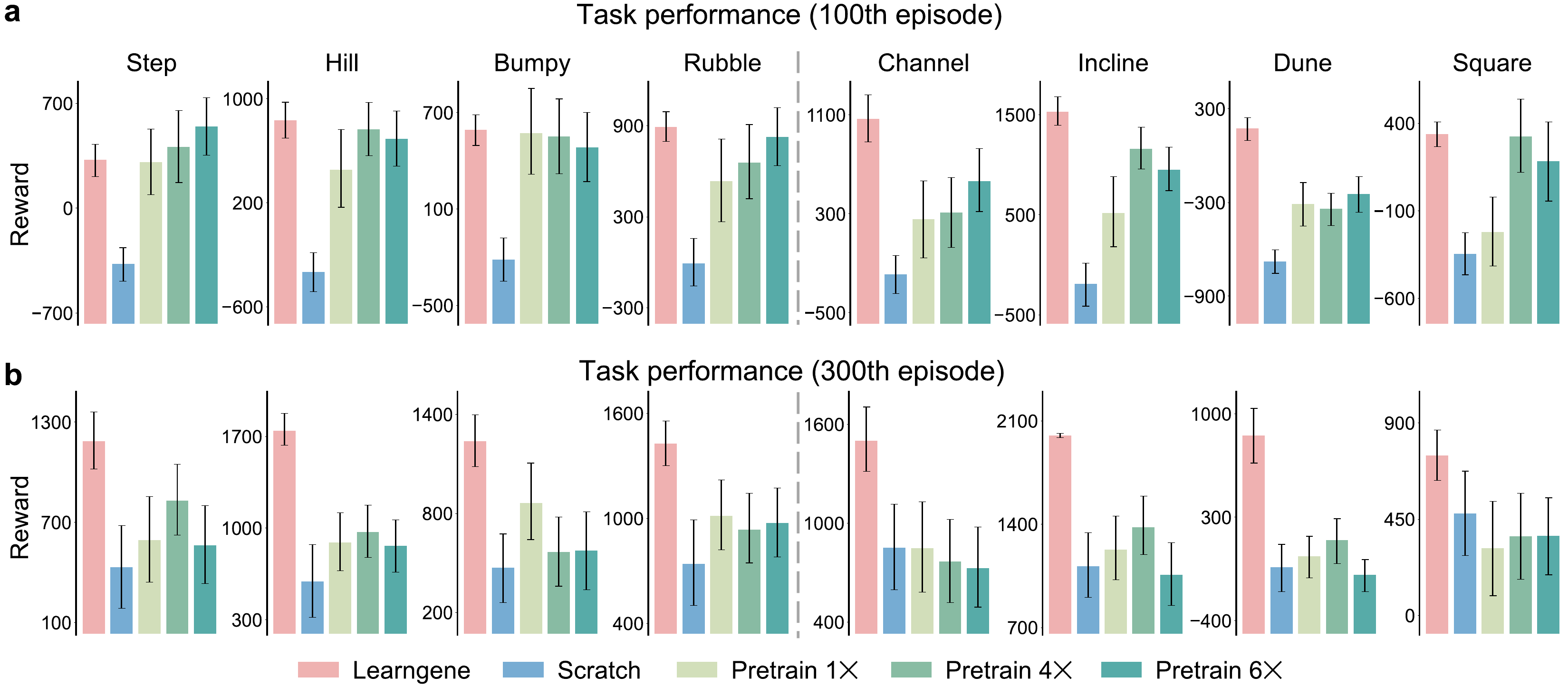}
  \caption{\textbf{Compared with pre-training and fine-tuning.} The bars indicate the average rewards ($n = 50$) of \textbf{(a)} the early training stages (i.e., the 100th episode) and \textbf{(b)} the later training stages (i.e., the 300th episode), with the error bars denoting the 95\% bootstrapped confidence intervals. Pretrain $i\times$ denotes the agents which are pre-trained with a lifetime $i$ times of that of the ordinary agents (i.e., the agents evolving in GRL, whose lifetime is 50 episodes).}
  \label{fig:pretrain and finetune}
\end{figure}

Learngenes can bring the agents instincts and better learning abilities through transferring the knowledge of their ancestors. However, that can also be achieved by the pre-trained models, one of the current research directions of transfer learning\cite{zhuang2020comprehensive,pratt1992discriminability, zoph2020rethinking}, which are trained on the large-scale datasets and then fine-tuned on the downstream tasks. Here, we obtain the pre-trained agents by training them in a terrain task combined with four obstacles (Fig. \ref{fig:8 obstacle}a), so that these agents can acquire the knowledge for four different obstacles simultaneously. Then, they are placed on the new tasks with all knowledge transferred (i.e., all parameters of the actor and critic networks).  

The pre-trained agents transfer all network parameters to the new tasks but fail to exhibit strong domain adaptation capabilities, because they rely on the transferred networks to exploit and have not efficiently explored the new terrains. What the pre-trained agents transfer is not only the common knowledge among the four training tasks (Fig. \ref{fig:8 obstacle}a), but also the task-specific knowledge for each task. With such knowledge, the agents can still gain benefits when they learn to control the coordination of their limbs to keep balance and avoid tumbling. Thus, in the early training stages, the pre-trained agents demonstrate strong environmental adaptability (Fig. \ref{fig:pretrain and finetune}a).

However, due to the significant differences between the new obstacles (Fig. \ref{fig:8 obstacle}b) and the training obstacles (Fig. \ref{fig:8 obstacle}a), the task-specific knowledge of the pre-trained agents hardly works, and even has a negative effect (i.e., negative transfer\cite{rosenstein2005transfer, wang2019characterizing}) when the agents start to walk forward and acquire the knowledge from the obstacles. The agents pre-trained for a longer period (with more task-specific knowledge) may not even be as good as those pre-trained with a shorter time, which is more obvious in the late training stages (Fig. \ref{fig:pretrain and finetune}b). Especially on the tasks with the Channel and Square obstacles, which the pre-trained agents have never experienced, they perform even worse than those trained from scratch (Fig. \ref{fig:pretrain and finetune}b).

In contrast, the agents inheriting the learngenes significantly outperform the pre-trained ones. The learngenes continuously accumulate the common knowledge of the training tasks in the inheritance of different agents with different tasks from generation to generation. 
Therefore, the agents inheriting the learngenes not only have the common knowledge (from the learngenes) among tasks, but also have the flexibility (from the non-learngene parts, which are initialized randomly) to learn new knowledge, thus maintaining their superiority on both the training obstacles and new obstacles (Fig \ref{fig:condense_gene}a,b).

\subsection{Learngenes condense common knowledge through the ``genomic bottleneck''.}
\begin{figure}[tb]
  \centering
  \includegraphics[width=39em]{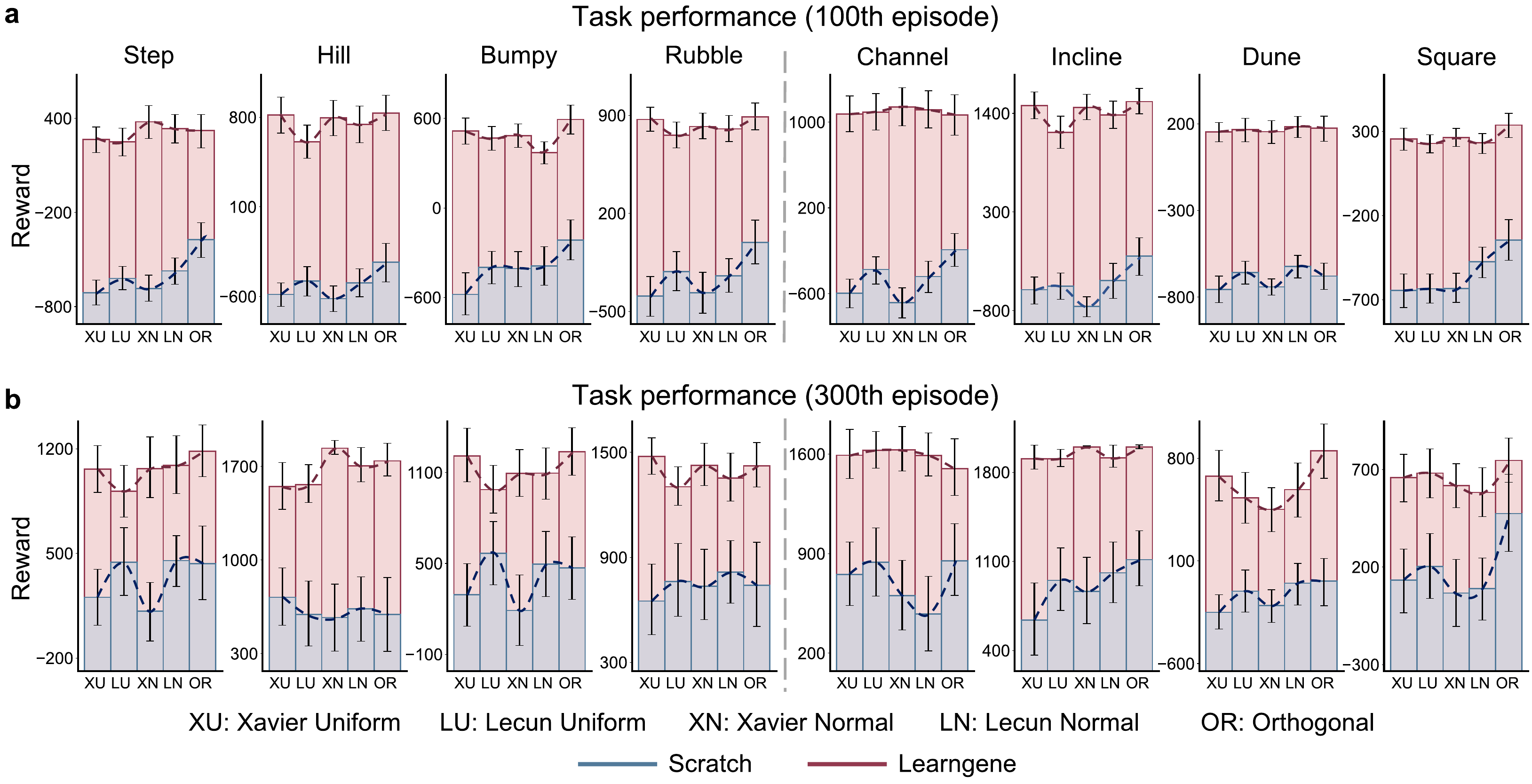}
  \caption{\textbf{Training with different initialization methods.} We choose five common initialization methods to initialize the non-learngene parts of the agents inheriting the learngenes and the agents learning from scratch. The bars indicate the average rewards ($n = 50$) of \textbf{(a)} the early training stages (i.e., the 100th episode) and \textbf{(b)} the later training stages (i.e., the 300th episode), with the error bars denoting the 95\% bootstrapped confidence intervals. The agents inheriting the learngenes tend to have similar learning abilities no matter how the non-learngene parts are initialized, but the agents from scratch are sensitive to the initialization methods.}
  \label{fig:initialization}
\end{figure}

As mentioned earlier, the nervous system of a newborn is not connected randomly, but wired up under the guidance of a genetic blueprint, which is compressed into the genes through the ``genomic bottleneck''\cite{zador2019critique}. 
Similarly, the knowledge of the agents is also condensed into their learngenes through such a ``genomic bottleneck'', which is essentially the common knowledge during the generational evolution of the agents.

The learngenes can ensure a preordained initialization of the agents' neural networks to a certain extent. Indeed, such initialization is actually the minimum necessary initialization.
Regardless of the initialization methods of the non-learngene components, the learngenes consistently facilitate the agents to attain almost equivalent environmental adaptability (Fig. \ref{fig:initialization}).
In contrast, the agents learning from scratch are susceptible to the initialization methods. Their abilities to learn and adapt largely depend on how the parameters of the neural networks are initialized. Generally, the orthogonal initialization\cite{saxe2014exact} seems to be the most advantageous method among all artificial mechanisms (Fig. \ref{fig:initialization}). 

\begin{figure}[tb]
  \centering
  \includegraphics[width=39em]{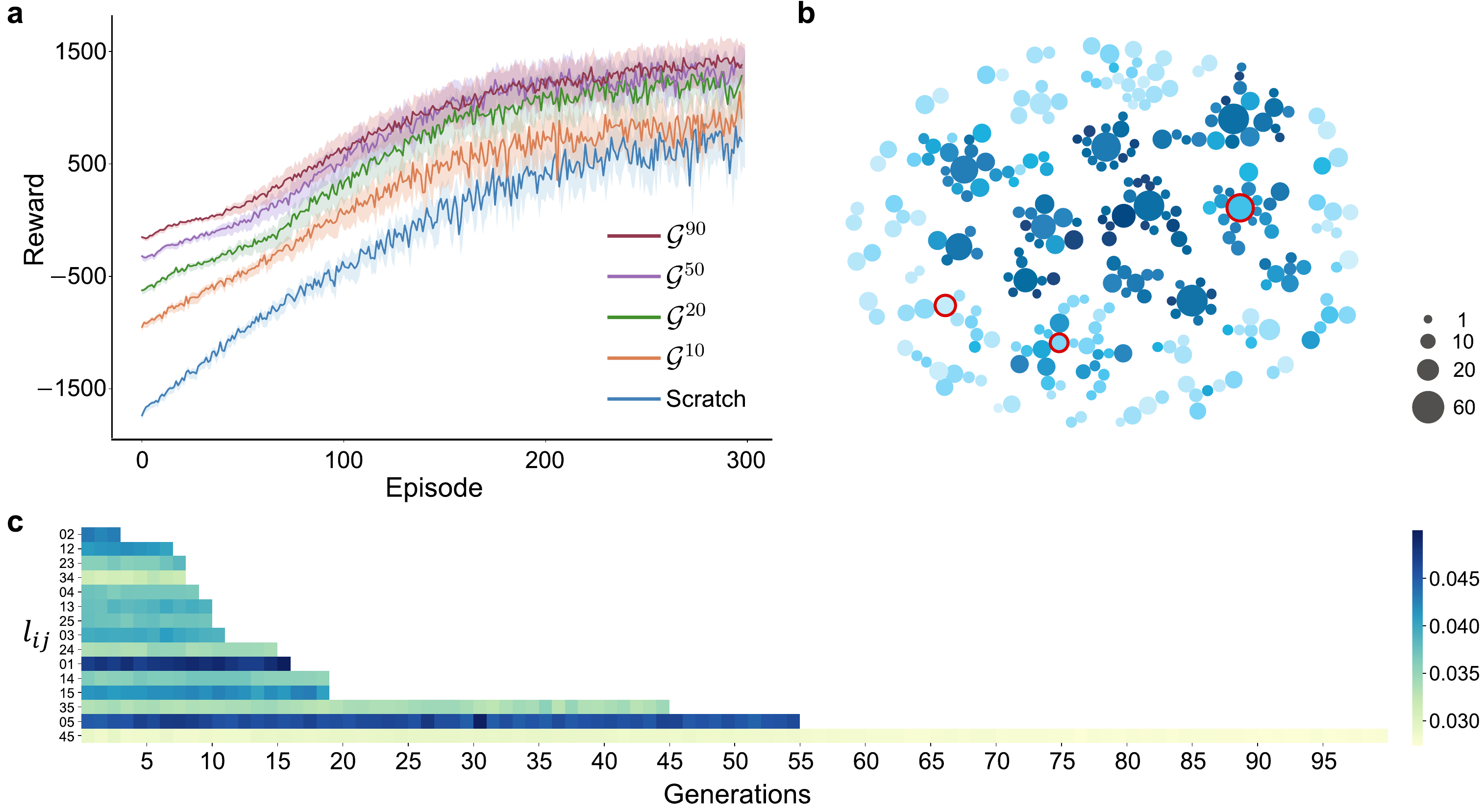}
  \caption{\textbf{Learngenes evolve with the evolution of the population.} \textbf{a} The training curves with the mean and 95\% bootstrapped confidence intervals of the rewards of the agents ($n=50$) inheriting the learngenes from the Gene Pool of the $i$th generation (i.e., $\mathcal{G}^i$). \textbf{b} The phylogenetic tree of the evolution (i.e., $(a, 2)$ in section \nameref{section:sec_gene_GEL}), where each dot represents a learngene in the Gene Pool, the dot size reflects the number of descendants, and the dot opacity reflects the fitness (darker indicates higher fitness). The tree demonstrates that the learngenes with higher fitness can be continuously evolved from the ancestral learngenes with lower fitness (i.e., the larger lighter dots selected by the red circle). \textbf{c} The heat map that records the parameter changes of the candidate learngenes $l_{ij}$ during evolution. The length of the heat map corresponding to $l_{ij}$ represents its generations surviving in evolution. The darker color represents that $l_{ij}$ have larger changes of the average parameter in this generation.}
  \label{fig:gene evolution}
\end{figure}

\subsection{Learngenes evolve through accumulating common knowledge.}
Darwinian evolution is a favored evolution of organisms in nature, which describes evolution as accumulating of small dominant mutations\cite{kimura1983neutral, jablonka1998lamarckian}. These small dominant mutations of genes create today's biological intelligence after accumulating around 3.5 billion years\cite{braga2017emperor, oro2004evolution}.

However, the evolution of the agents shows a different pattern from the biological evolution in nature, which strongly demonstrates the Lamarckian inheritance.
As an early theory of evolution, Lamarckism considered that what individuals acquired in physiology during their life can be transmitted to their descendants, who would have instincts soon after birth\cite{larson2004evolution, gould2010lying}. 

The agents born with the learngenes, which can be seen as the ``memory'' of the entire population, show strong instincts (Fig. \ref{fig:instinct}). The agents continuously gain the experience and knowledge by interacting with the environments, which are encoded into the learngenes in the form of updating weights of the neural networks, thus promoting the continuous evolution of the learngenes during their lifetime. 
The evolved learngenes will be passed on to the descendants through the inheritance, bring the experience and knowledge of the ancestors, and begin the encoding of the experience in a new generation. 
As a result, the learngenes evolving with more generations encode more knowledge of the population, which bring higher fitness to the agents inheriting them at birth. 
For example, the fitness of the newborn agents inheriting the learngenes from the $90^{th}$ generation even excels that of the agents learning from scratch after half of their lifetime (Fig. \ref{fig:gene evolution}a). 
By quickly turning more advanced ancestry experiences into instincts, the agents will adapt to the environments faster and better through their learning, continuously increasing the ceiling of the population's adaptability to the environments (Fig. \ref{fig:gene evolution}a). 

\subsection{Learngenes exhibit the diversity and continuity during evolution.}
Genetic diversity provides the raw material for evolution\cite{hughes2008ecological}, and learngenes also demonstrate such diversity. The early learngenes with low fitness can also evolve into the better ones and produce a large number of superior descendants by accumulating common knowledge during evolution (Fig. \ref{fig:gene evolution}b). 
That's why GRL did not directly eliminate learngenes with relative low fitness, but also give them the opportunity to reproduce (see section \nameref{sec:sec_genepool} and section \nameref{sec:sec_inherit_gene}).

The gene in nature is a relatively stable genetic unit, which can be stably inherited across generations\cite{watson1953molecular, hochschild1986lambda, baker2008molecular}. In the same vein, the learngenes also possess such continuity during the generational evolution of the intelligent agents. Learngenes always maintain a relative high probability of generating descendants during the whole evolution, and achieve stable inheritance in the later stage (Fig. \ref{fig:condense_gene}c). Also, we quantify the average changes of the learngenes during evolution by measuring the alteration of the candidate learngenes' parameters in the Manhattan distance during their lifetime. 
The parameters of the learngenes have small changes during the evolution process, and exhibit more and more stability during evolution (Fig. \ref{fig:gene evolution}c). 

\section*{Discussion}
In nature, genes accumulate the knowledge of ancestors through evolution, and pass it to descendants through inheritance. Gradually, animals exhibit some innate behaviors at birth, which significantly help them better adapt to their living environments. 
However, the current AI systems usually learn from scratch (without genes), which causes significant difference between the machine intelligence and the biological intelligence.
In this work, we define the learngenes in the intelligent agents and train them with the inheritance of the learngenes via Genetic Reinforcement Learning (GRL), a computational framework for learning and evolution. 
The GRL represents the learngenes by the fragments of networks, evaluates the learngenes by the fitness of the agents in RL, and evolves the learngenes across generations.

We represent the learngenes in the agents as the combination of the last two layers from the actor network. The learngenes have at least the following advantages. First, the learngenes bring instincts and strong learning abilities to the agents. 
The agents inheriting the learngenes move forward unconsciously and outperform those learning from scratch during the whole lifetime. 
Second, the learngenes transfer common knowledge to the agents from their ancestors.
The agents inheriting the learngenes beat the pre-trained agents, implying that the knowledge condensed in the learngenes is the common knowledge among various tasks. In comparison, the knowledge in the pre-train agents is more correlated to the training tasks.
Third, the learngenes condense common knowledge through the “genomic bottleneck”. As the learngenes have already initialized the core parts of the ``brains'' of the agents to some extent, the agents inheriting the learngenes are insensitive to the initialization of the non-learngene parts. 

We reveal that the evolution of the intelligent agents demonstrates the Lamarckian inheritance. The learngenes evolve continuously and stably with the evolution of the population, as the agents steadily encode the acquired knowledge into the learngenes and pass them on to the next generation. In this process, the learngenes exhibit the diversity and continuity.

Overall, we expect that the proposal of the learngenes in the intelligent agents can bring some novel insights to the AI research and take the machine intelligence one more step toward the biological intelligence. 

\begin{methods}
\label{section:sec_method}

\subsection{Generational Learning and Evolution.}
We propose a computational framework called Genetic Reinforcement Learning (GRL) to simultaneously train the agents and evolve the learngenes through multiple generations (Fig. \ref{fig:GRL}). Each generation starts with a population of $n_p$ agents. Each agent inherits the learngenes from previous generations and is randomly assigned a task from $m$ tasks. 
The evolution starts after the lifetime learning of all $n_p$ agents. Specifically, every $s$ agents are randomly selected (without replacement) to engage in a tournament. In each tournament, the candidate learngenes of the winners have a chance to stay in the Gene Pool for the following generations. 
After all tournaments have finished, a new generation runs in a nested cycle of learning and evolution. 

\subsection{Reinforcement Learning.}
\label{section:sec_life_time_learning}
GRL trains the agents using a RL style, where the agents learn experience and knowledge by interacting with the environments. 
We use the MuJoCo simulator\cite{todorov2012mujoco} to conduct the experiments. The agent in our experiments is the ant robot with nine parts, including a torso and four legs (each leg has two links), which are connected by eight hinges. By applying torques on the hinges, the ant robot can coordinate its four legs to move in a certain direction.

Each agent is controlled by an actor network and a critic network\cite{konda1999actor}, which both have five hidden layers with 48 neurons. The inputs of the actor and critic network are the proprioceptive observations (27 dimensions) of the agent at each time step, including the coordinate, orientation, joint angles, coordinate velocity and angular velocities, which are provided in the MuJoCo simulator\cite{todorov2012mujoco}. The outputs of the actor network are the torques applied at the hinge joints (8 dimensions) of the agent, which controls its actions.

We use the PPO algorithm\cite{schulman2017proximal} to optimize the parameters. For all terrain tasks, the moving velocity component of the agents in the direction toward the finishing line is the only source of positive rewards. At each time step $t$, the received rewards $r_t$ is
\begin{equation}
    r_t = \gamma v_{\text{target}} - \delta ||o_{\text{action}}||^2
\label{equ:equ_reward}
\end{equation}
where $v_{\text{target}}$ is the component of the velocity toward the finishing line, $o_{\text{action}}$ is the action taken by the agents, and $\gamma$ and $\delta$ are the balance weights. The second term of the above equation penalizes too large actions. 
Besides, the rewards $r_e$ of an agent in each episode $e$ is the sum of $r_t$ about the whole time steps ($t_\text{end}$) needed in $e$ (the maximum $t_\text{end}$ is 3000), which is calculated as 
\begin{equation}
    r_e = \sum_{t=0}^{t_\text{end}} r_t
\label{equ:equ_reward_epi}
\end{equation}
Note that the ``rewards'' in this paper all refer to $r_e$ if not otherwise specified. 

\subsection{Environments.}
GRL provides $m=4$ task environments to train the agents. Each one consists of a rectangle arena of size 120$\times$40 square meters ($\text{m}^2$) with one type of obstacle from Step, Bumpy, Hill, and Rubble (Fig. \ref{fig:8 obstacle}a). In each generation, each agent will randomly select an environment as its own task room, where more than half of the room is filled with obstacles, with the remaining flat (Fig. \ref{fig:8 obstacle}c). A newborn agent starts from the starting line, tries to cross the obstacles, and reaches the finishing line. 

In the forward direction toward the finish line, the difficulty of the obstacles gradually increases in terms of the height and slope of the obstacles. \textbf{Step} has a height in the range $[\text{0.10}, \text{0.13}]$m and length in the range $[\text{2}, \text{3}]$m. A step sequence is $\tau-$steps up followed by $\tau-$steps down, where $\tau \in [\text{3}, \text{6}]$. \textbf{Bumpy} has a trapezoidal shape, with a height in the range $[-\text{0.20}, \text{0.24}]$m and a width of 2m. \textbf{Hill} is a semi-cylindrical structure constructed by sin wave, with a height in the range $[\text{0.15}, \text{0.60}]$m and a width in the range $[\text{4}, \text{10}]$m. \textbf{Rubble} is composed of a right square pyramid, with a height in the range $[\text{0.25}, \text{0.50}]$m and a width in the range $[\text{2}, \text{4}]$m. 

When setting up the tasks, we ensured as much as possible that the training obstacles used to extract the learngenes (i.e., Step, Bumpy, Hill and Rubble) have similar difficulties for the agents to cross. Meanwhile, in order to better extract the learngenes from the generational evolution of the agents, we made a certain balance between the similarity and difference among the training obstacles (Fig. \ref{fig:8 obstacle}a), which is convenient for the learngenes to condense the common knowledge. As for the new obstacles (i.e., Channel, Incline, Dune and Square) set for testing the performance of the learngenes, we choose to maximize the differentiation (both the difficulty of the tasks and the shape of the obstacles) from the other tasks to better demonstrate the common knowledge contained in the learngenes (Fig. \ref{fig:8 obstacle}b). 

To demonstrate the difference and similarity between the obstacles more intuitively, we introduced the concept of the knowledge transfer rate, which reflects the similarity of the knowledge obtained by the agents on different obstacles (the heat maps in Fig. \ref{fig:8 obstacle}). Concretely, the knowledge transfer rate $\mathcal{T}_{ij}$ from the task with the obstacle $t_j$ to the task with the obstacle $t_i$ (hereafter the task $t_j$ and $t_i$ respectively) is calculated as
\begin{equation}
    \mathcal{T}_{ij} = \frac{R_{ji}-w_i}{R_{ii}-w_i} 
\label{equ:knowledge_tranfer}
\end{equation}
where $R_{ji}$ is the rewards on the task $t_i$ obtained by the agents which have trained on the task $t_j$ (transfer the agents directly without training). $w_i$ is the rewards on the task $t_i$ when the agents successfully stand and attempt to walk (the agents begin to acquire the knowledge from the obstacles at this point). In Fig. \ref{fig:8 obstacle}, we train the agents ($n=50$) from scratch on the task $t_i$ with 300 episodes, and $w_i$ is the rewards on the 150th episode.  

\subsection{Tournaments.}
Specifically, let $P=\{\alpha_1, \alpha_2, \cdots, \alpha_{n_p}\}$ ($n_p=50$) be a population of agents. 
After all the agents in $P$ have completed their lifetime learning, GRL conducts the tournaments. 

For each agent $\alpha_i$, GRL adopts the average of $r_e$ as the fitness $f_i$ (a constant $\zeta=1000$ is added to $f_i$ to ensure positive fitness in our experiments) to evaluate its performance throughout the lifetime, 
\begin{equation}
    f_i = \frac{\sum_{e=0}^{lt} r_e}{lt} + \zeta
\label{equ:fitness}
\end{equation}
where $lt$ is the lifetime of an agent (i.e., the number of episodes that an agent can complete and $lt=50$ in the experiments). The fitness $f_i$ indicates the learning ability and the environmental adaptability of $\alpha_i$. 

To ensure the fairness of the tournaments across different tasks, the fitness of the agents will be normalized according to the difficulty of the task, which is determined by the fitness of all the agents completing this task. Concretely, for an agent $\alpha_i$ with the fitness $f_i$ on task $t$, the normalized fitness $\tilde{f}_i$ is calculated by
\begin{equation}
    \tilde{f}_i = \frac{f_i - \min\limits_{f \in F^{(t)}} f}{\max\limits_{f \in F^{(t)}} f-\min\limits_{f \in F^{(t)}} f}\cdot \frac{\sum\limits_{f \in F}f}{n}
\label{equ:equ_fitness_normal}
\end{equation}
where the set $F$ contains the fitness of all the agents in $P$, and $F^{(t)}$ is the subset of $F$ and consists of the fitness of the agents with task $t$. That is, the normalized fitness is calculated by a min-max normalization on the current task and then multiplying the mean fitness of all the agents. 

GRL randomly selects $s$ ($s=3$) agents from $P$ (without replacement) to engage in a tournament. The sub-population of the winners is denoted by $P^*=\{\alpha_1^*, \alpha_2^*, ..., \alpha_c^*\}$ ($c=17$ with $n_p=50$ and $s=3$). The sub-population of the winners $P^*$ consists of the agents with better learning abilities. Therefore, it likely contains better candidate learngenes. 

\subsection{Learngene.}
\label{section:sec_learngene}
As introduced above, each agent adopts the actor-critic algorithm, which can be viewed as a stochastic gradient algorithm on the parameter space of the actor network\cite{konda1999actor}. The learngenes of the agents are modeled as the combination of layers from the actor or critic networks. In nature, due to the limited capacity of the genome, the innate processes of the organisms captured by evolution need to be compressed into the genome through a ``genomic bottleneck''\cite{zador2019critique}. Similarly, as for the learngenes in the agents, we consider that there also exist a ``bottleneck'' to compress common knowledge of the agents. As a result, the learngenes are composed of up to five layers from the actor or critic networks (instead of the whole actor or critic networks).

We have found in the experiments (section \nameref{section:sec_gene_GEL}) that the critic networks have no common knowledge which can be transmitted to the descendants. And the common knowledge is concentrated in the two layers of the actor networks (Fig. \ref{fig:condense_gene}a,b), so the learngenes in the intelligent agents take the form of two layers from the actor network (i.e., $(a,2)$).

Next, GRL selects the candidate learngenes from $P^*$ for evolution in the following generations. To ensure that the learngenes have the continuity and diversity like the genes in nature\cite{brandt2013ancient, achtman2008microbial}, we introduced two essential parts when selecting the candidate learngenes, including the \textbf{Gene Pool} and \textbf{Gene Tree}, which store the superior ancestral candidate learngenes and the kinship of the candidate learngenes throughout the whole evolution, respectively.

\subsection{Gene Pool.}
\label{sec:sec_genepool}
We define a new data structure called Gene Pool (GP) to store the superior candidate learngenes (candidate solutions) for evolution. 
Since the learngenes take the form of a combination of layers, there would be $C_6^{\kappa}$ combinations in the GP if each one consists of $\kappa$ layers. Specifically, for $\kappa = 2$, $\text{GP} = \{l_{01}^{(a)}, l_{02}^{(a)}, l_{03}^{(a)}, ...,l_{ij}^{(a)},..., l_{45}^{(a)}\}$, where $l_{ij}^{(a)}$ represents the combination of the $i$th and $j$th layers of the actor network. 
At the initial generation, each candidate form of the learngene in the GP has $\rho_{\max}=7$ candidate learngenes (i.e., $\text{GP}(l_{01}) = \{l_{01}(\hat{\alpha}_1), l_{01}(\hat{\alpha}_2), ..., l_{01}(\hat{\alpha}_{\rho_{\max}})\}$), which are randomly extracted from $\hat{\alpha}_i$ in $P^*$. So, there exists $C_6^{\kappa}\cdot \rho_{\max}$ learngenes in the GP (Supplementary Fig. 2). 

Each candidate learngene $l_{ij}(\cdot)$ in the GP has a score $s_g$, which represents its quality. The score $s_{g}$ is determined by both the fitness of the winner agent where $l_{ij}(\cdot)$ is extracted from and the effective layer width (defined in Eq. (\ref{equ:equ_score_gene})) of the learngene form that makes up $l_{ij}(\cdot)$. Specifically, for a candidate learngene $l_{ij}(\alpha)$ which is extracted from $\alpha$ with the fitness $f$, its score is calculated by
\begin{equation}
    s_{g} = \frac{f}{\sum\limits_{\theta \in \{i,j\}} \text{elw}(l_{\theta})}
\label{equ:equ_score_gene}
\end{equation}
where $\text{elw}(l) = \sqrt{|l|}$ calculates the effective layer width of the layer $l$, and $|\cdot|$ represents the parameter amount of one layer. In the following generations, we will select the superior candidate learngenes from $P^*$, and replace the candidate learngenes with lower scores in the GP. To prevent significant changes in the GP and ensure that the better candidate learngenes in the GP have the opportunity to live for more generations, at most two candidate learngenes can be replaced for each candidate form of the learngene.

\subsection{Gene Tree.}
We also define a new data structure called Gene Tree (GT) to store the kinship of the candidate learngenes during the entire evolution process. 
The candidate learngenes with the same form in the GP make up a gene forest, and the candidate learngenes in such forest make up the GT based on their blood relationships. The root nodes of each GT are the candidate learngenes in the GP at the initial generation. In each generation, the candidate learngene extracted from a winner $\alpha^*$ will also be added to the GT as a leaf node, whose parent node is the candidate learngene inherited by $\alpha^*$.
So, there exist $C_6^{\kappa}$ gene forests and $C_6^{\kappa}\cdot \rho_{\max}$ gene trees in total. The path length between two nodes represents the closeness of the parental generation between the corresponding candidate learngenes.  

After the tournaments, new candidate learngenes are extracted from the winners. After calculating the scores of the new candidate learngenes as Eq. (\ref{equ:equ_score_gene}), we also update the scores of the winners' ancestral candidate learngenes according to the GT to maintain their excellence and continuity (Supplementary Fig. 3). 
It will start from a leaf node (i.e., the candidate learngene) that a winner inherits, and backtrack to the root node. 
The scores of the ancestry candidate learngenes (in the GP) in this update path are updated as
\begin{equation}
    s_{g_a} = s_{g_a} + \text{{sim}}(g_a, g_d) \eta^{l+1} f
\label{equ:equ_update_score}
\end{equation}
where both $g_a$ and $g_d$ are the ancestry candidate learngenes, and $g_a$ is the father of $g_d$. $\eta$ is the parental decay coefficient, and $l$ is the path length between $g_a$ and the leaf node on the GT. $f$ is the fitness of an agent winning the tournament. 
The function $\text{sim}(\cdot, \cdot)$ represents the learngene similarity, which is calculated as
\begin{equation}
    \text{{sim}}(g_a, g_d) = \frac{\sum\limits_{l \in \mathcal{L}_{g_a} \cap \mathcal{L}_{g_d}} \text{elw}(l)}{\sum\limits_{l^{'} \in \mathcal{L}_{g_a} \cup \mathcal{L}_{g_d}} \text{elw}(l^{'})}
\label{equ:equ_gene_similarity}
\end{equation}
where $\mathcal{L}_{g_a}$ and $\mathcal{L}_{g_d}$ are the sets of layers comprising $g_a$ and $g_d$, respectively. Specifically, for a candidate learngene $l_{01}(\cdot)$, $\mathcal{L}_{g}$ is $\{l_0, l_1\}$.

Just as the loss of genes may promote the evolution of organisms and the diversity of life in nature\cite{guijarro2020widespread, sharma2018genomics, xue2023functional}, the scores of the candidate learngenes in the entire GP will undergo a certain proportion of decay $\beta$ after each generation to accelerate the elimination of the early candidate learngenes, thus promoting the continuous evolution and diversity of the learngenes.

\subsection{Extract Learngenes.}
\label{sec:sec_condense_gene}
After GT and GP are constructed, we have a complete learngene evaluation and storage mechanism. Next, we try to extract the superior candidate learngenes from $P^*$. In order that the candidate learngene forms with low scores also have a chance to produce candidate learngenes, we introduce randomness when extracting the candidate learngenes during evolution. To achieve that, we define the form probability (Supplementary Fig. 2), which represents the probability of one candidate becoming the learngene form. Concretely, the form probability of $l_{ij}$ is calculated by 
\begin{equation}
    p_{l_{ij}} = \frac{s_{l_{ij}}}{\sum_{l\in \text{GP}} s_{l}}
\label{equ:equ_layer_probability}
\end{equation}
where $s_{l_{ij}}$ is the score of $l_{ij}$. It is determined by the score of all candidate learngenes (in the GP of the current generation) of this candidate form: 
\begin{equation}
    s_{l_{ij}} = \sum_{g\in l_{ij}} s_{g}
\end{equation}

We will extract the candidate learngenes based on the form probability to ensure that GRL has a larger search space for learngenes (i.e., the diversity of the learngenes). Meanwhile, considering the continuity of the paternal learngenes (i.e., the candidate learngene $l_p(\cdot)$ inherited by a winner), the probability of extracting the candidate learngenes is calculated by
\begin{equation}
    \mathcal{P}_{l_{ij}} = \frac{h_{l_{ij}}}{\sum_{l\in \text{GP}} h_{l}}
\end{equation}
where
\begin{equation}
    h_{l_{ij}} = \begin{cases}p_{l_{ij}},\quad &l_{ij}\neq l_{p}\\1,\quad &l_{ij}=l_{p}\end{cases}
\end{equation}
Finally, we place the extracted candidate learngenes in the corresponding positions of the GP. 

\subsection{Inherit Learngenes.}
\label{sec:sec_inherit_gene}
Next, we will choose the candidate learngenes from the GP and give them the opportunity to reproduce, that is, the learngenes are passed on to the next generation.
When a new generation of agents is born, each agent will randomly select a candidate learngene from the GP according to the form probability and the score of each candidate learngene, which makes the candidate learngenes with lower scores in the GP still have the opportunity to reproduce, thus maintaining the diversity of the learngenes.
The probability of a candidate learngene $l_{ij}(\cdot)$ (in the GP) being selected is calculated by
\begin{equation}
    p_{g} = p_{l_{ij}}\cdot \frac{s_g}{s_{l_{ij}}}
\end{equation}
Then, the candidate learngenes will be copied to the corresponding network layers of the agents, and the non-learngene parts will be initialized randomly (Fig. \ref{fig:GRL}e). 
After all the agents have inherited the candidate learngenes, they will start the lifetime RL (section \nameref{section:sec_life_time_learning}) and a new round of evolution.

\end{methods}

\section*{Data availability}
The configuration files necessary to reproduce the data used in this work have been made available on Github (\href{https://github.com/fu-feng/GRL}{https://github.com/fu-feng/GRL}).

\section*{Code availability}
The Python implementation of GRL has been released on GitHub (\href{https://github.com/fu-feng/GRL}{https://github.com/fu-feng/GRL}) with the code, setup instructions, and configuration files needed for reproducing the results of the paper.
\section*{References}
\vspace{1em}
\bibliography{reference.bib}

\begin{thebibliography}{10}
\expandafter\ifx\csname url\endcsname\relax
  \def\url#1{\texttt{#1}}\fi
\expandafter\ifx\csname urlprefix\endcsname\relax\def\urlprefix{URL }\fi
\providecommand{\bibinfo}[2]{#2}
\providecommand{\eprint}[2][]{\url{#2}}

\bibitem{braga2017emperor}
\bibinfo{author}{Braga, A.} \& \bibinfo{author}{Logan, R.~K.}
\newblock \bibinfo{title}{{The emperor of strong AI has no clothes: limits to
  artificial intelligence}}.
\newblock \emph{\bibinfo{journal}{Information}} \textbf{\bibinfo{volume}{8}},
  \bibinfo{pages}{156} (\bibinfo{year}{2017}).

\bibitem{oro2004evolution}
\bibinfo{author}{Or{\'o}, J.~J.}
\newblock \bibinfo{title}{Evolution of the brain: from behavior to
  consciousness in 3.4 billion years}.
\newblock \emph{\bibinfo{journal}{Neurosurgery}} \textbf{\bibinfo{volume}{54}},
  \bibinfo{pages}{1287--1297} (\bibinfo{year}{2004}).

\bibitem{danchin2018cultural}
\bibinfo{author}{Danchin, E.} \emph{et~al.}
\newblock \bibinfo{title}{Cultural flies: Conformist social learning in
  fruitflies predicts long-lasting mate-choice traditions}.
\newblock \emph{\bibinfo{journal}{Science}} \textbf{\bibinfo{volume}{362}},
  \bibinfo{pages}{1025--1030} (\bibinfo{year}{2018}).

\bibitem{howard2019numerical}
\bibinfo{author}{Howard, S.~R.}, \bibinfo{author}{Avargu{\`e}s-Weber, A.},
  \bibinfo{author}{Garcia, J.~E.}, \bibinfo{author}{Greentree, A.~D.} \&
  \bibinfo{author}{Dyer, A.~G.}
\newblock \bibinfo{title}{Numerical cognition in honeybees enables addition and
  subtraction}.
\newblock \emph{\bibinfo{journal}{Sci. Adv.}} \textbf{\bibinfo{volume}{5}},
  \bibinfo{pages}{eaav0961} (\bibinfo{year}{2019}).

\bibitem{hassabis2017neuroscience}
\bibinfo{author}{Hassabis, D.}, \bibinfo{author}{Kumaran, D.},
  \bibinfo{author}{Summerfield, C.} \& \bibinfo{author}{Botvinick, M.}
\newblock \bibinfo{title}{Neuroscience-inspired artificial intelligence}.
\newblock \emph{\bibinfo{journal}{Neuron}} \textbf{\bibinfo{volume}{95}},
  \bibinfo{pages}{245--258} (\bibinfo{year}{2017}).

\bibitem{matsuo2022deep}
\bibinfo{author}{Matsuo, Y.} \emph{et~al.}
\newblock \bibinfo{title}{Deep learning, reinforcement learning, and world
  models}.
\newblock \emph{\bibinfo{journal}{Neural Netw.}}
  \textbf{\bibinfo{volume}{152}}, \bibinfo{pages}{267--275}
  (\bibinfo{year}{2022}).

\bibitem{kriegeskorte2015deep}
\bibinfo{author}{Kriegeskorte, N.}
\newblock \bibinfo{title}{Deep neural networks: a new framework for modeling
  biological vision and brain information processing}.
\newblock \emph{\bibinfo{journal}{Annu. Rev. Vis. Sci.}}
  \textbf{\bibinfo{volume}{1}}, \bibinfo{pages}{417--446}
  (\bibinfo{year}{2015}).

\bibitem{sutton2018reinforcement}
\bibinfo{author}{Sutton, R.~S.} \& \bibinfo{author}{Barto, A.~G.}
\newblock \emph{\bibinfo{title}{Reinforcement learning: An introduction}}
  (\bibinfo{publisher}{MIT press}, \bibinfo{year}{2018}).

\bibitem{gronauer2022multi}
\bibinfo{author}{Gronauer, S.} \& \bibinfo{author}{Diepold, K.}
\newblock \bibinfo{title}{Multi-agent deep reinforcement learning: a survey}.
\newblock \emph{\bibinfo{journal}{Artif. Intell. Rev.}}
  \textbf{\bibinfo{volume}{55}}, \bibinfo{pages}{895–943}
  (\bibinfo{year}{2022}).

\bibitem{nguyen2020deep}
\bibinfo{author}{Nguyen, T.~T.}, \bibinfo{author}{Nguyen, N.~D.} \&
  \bibinfo{author}{Nahavandi, S.}
\newblock \bibinfo{title}{Deep reinforcement learning for multiagent systems: A
  review of challenges, solutions, and applications}.
\newblock \emph{\bibinfo{journal}{IEEE Trans. Cybern.}}
  \textbf{\bibinfo{volume}{50}}, \bibinfo{pages}{3826--3839}
  (\bibinfo{year}{2020}).

\bibitem{silver2016mastering}
\bibinfo{author}{Silver, D.} \emph{et~al.}
\newblock \bibinfo{title}{Mastering the game of go with deep neural networks
  and tree search}.
\newblock \emph{\bibinfo{journal}{Nature}} \textbf{\bibinfo{volume}{529}},
  \bibinfo{pages}{484--489} (\bibinfo{year}{2016}).

\bibitem{silver2017mastering}
\bibinfo{author}{Silver, D.} \emph{et~al.}
\newblock \bibinfo{title}{Mastering the game of go without human knowledge}.
\newblock \emph{\bibinfo{journal}{Nature}} \textbf{\bibinfo{volume}{550}},
  \bibinfo{pages}{354--359} (\bibinfo{year}{2017}).

\bibitem{van2023chatgpt}
\bibinfo{author}{Van~Dis, E.~A.}, \bibinfo{author}{Bollen, J.},
  \bibinfo{author}{Zuidema, W.}, \bibinfo{author}{van Rooij, R.} \&
  \bibinfo{author}{Bockting, C.~L.}
\newblock \bibinfo{title}{{ChatGPT: five priorities for research}}.
\newblock \emph{\bibinfo{journal}{Nature}} \textbf{\bibinfo{volume}{614}},
  \bibinfo{pages}{224--226} (\bibinfo{year}{2023}).

\bibitem{abio2023ai}
\bibinfo{author}{ABIO, B.}
\newblock \bibinfo{title}{{In AI, is bigger better?}}
\newblock \emph{\bibinfo{journal}{Nature}} \textbf{\bibinfo{volume}{615}},
  \bibinfo{pages}{202--205} (\bibinfo{year}{2023}).

\bibitem{landhuis2017neuroscience}
\bibinfo{author}{Landhuis, E.}
\newblock \bibinfo{title}{{Neuroscience: Big brain, big data}}.
\newblock \emph{\bibinfo{journal}{Nature}} \textbf{\bibinfo{volume}{541}},
  \bibinfo{pages}{559--561} (\bibinfo{year}{2017}).

\bibitem{krink1997analysing}
\bibinfo{author}{Krink, T.} \& \bibinfo{author}{Vollrath, F.}
\newblock \bibinfo{title}{Analysing spider web-building behaviour with
  rule-based simulations and genetic algorithms}.
\newblock \emph{\bibinfo{journal}{J. Theor. Biol.}}
  \textbf{\bibinfo{volume}{185}}, \bibinfo{pages}{321--331}
  (\bibinfo{year}{1997}).

\bibitem{gorissen2017development}
\bibinfo{author}{Gorissen, B.} \emph{et~al.}
\newblock \bibinfo{title}{The development of locomotor kinetics in the foal and
  the effect of osteochondrosis}.
\newblock \emph{\bibinfo{journal}{Equine Vet. J.}}
  \textbf{\bibinfo{volume}{49}}, \bibinfo{pages}{467--474}
  (\bibinfo{year}{2017}).

\bibitem{seung2012connectome}
\bibinfo{author}{Seung, S.}
\newblock \emph{\bibinfo{title}{Connectome: How the brain's wiring makes us who
  we are}} (\bibinfo{publisher}{HMH}, \bibinfo{year}{2012}).

\bibitem{wong2015behavioral}
\bibinfo{author}{Wong, B.~B.} \& \bibinfo{author}{Candolin, U.}
\newblock \bibinfo{title}{Behavioral responses to changing environments}.
\newblock \emph{\bibinfo{journal}{Behav. Ecol.}} \textbf{\bibinfo{volume}{26}},
  \bibinfo{pages}{665--673} (\bibinfo{year}{2015}).

\bibitem{sih2011evolution}
\bibinfo{author}{Sih, A.}, \bibinfo{author}{Ferrari, M.~C.} \&
  \bibinfo{author}{Harris, D.~J.}
\newblock \bibinfo{title}{Evolution and behavioural responses to human-induced
  rapid environmental change}.
\newblock \emph{\bibinfo{journal}{Evol. Appl.}} \textbf{\bibinfo{volume}{4}},
  \bibinfo{pages}{367--387} (\bibinfo{year}{2011}).

\bibitem{tan2022rlx2}
\bibinfo{author}{Tan, Y.}, \bibinfo{author}{Hu, P.}, \bibinfo{author}{Pan, L.},
  \bibinfo{author}{Huang, J.} \& \bibinfo{author}{Huang, L.}
\newblock \bibinfo{title}{Rlx2: Training a sparse deep reinforcement learning
  model from scratch}.
\newblock In \emph{\bibinfo{booktitle}{Proceedings of the International
  Conference on Learning Representations}} (\bibinfo{year}{2022}).

\bibitem{riedmiller2018learning}
\bibinfo{author}{Riedmiller, M.} \emph{et~al.}
\newblock \bibinfo{title}{Learning by playing solving sparse reward tasks from
  scratch}.
\newblock In \emph{\bibinfo{booktitle}{Proceedings of the International
  Conference on Machine Learning}}, \bibinfo{pages}{4344--4353}
  (\bibinfo{year}{2018}).

\bibitem{bakhtin2021no}
\bibinfo{author}{Bakhtin, A.}, \bibinfo{author}{Wu, D.},
  \bibinfo{author}{Lerer, A.} \& \bibinfo{author}{Brown, N.}
\newblock \bibinfo{title}{No-press diplomacy from scratch}.
\newblock \emph{\bibinfo{journal}{Proceedings of Advances in Neural Information
  Processing Systems}} \textbf{\bibinfo{volume}{34}},
  \bibinfo{pages}{18063--18074} (\bibinfo{year}{2021}).

\bibitem{agarwal2022reincarnating}
\bibinfo{author}{Agarwal, R.}, \bibinfo{author}{Schwarzer, M.},
  \bibinfo{author}{Castro, P.~S.}, \bibinfo{author}{Courville, A.~C.} \&
  \bibinfo{author}{Bellemare, M.}
\newblock \bibinfo{title}{Reincarnating reinforcement learning: Reusing prior
  computation to accelerate progress}.
\newblock In \emph{\bibinfo{booktitle}{Proceedings of Advances in Neural
  Information Processing Systems}}, \bibinfo{pages}{28955--28971}
  (\bibinfo{year}{2022}).

\bibitem{li2018a2}
\bibinfo{author}{Li, D.}, \bibinfo{author}{Wu, H.}, \bibinfo{author}{Zhang, J.}
  \& \bibinfo{author}{Huang, K.}
\newblock \bibinfo{title}{A2-rl: Aesthetics aware reinforcement learning for
  image cropping}.
\newblock In \emph{\bibinfo{booktitle}{Proceedings of the IEEE Conference on
  Computer Vision and Pattern Recognition}}, \bibinfo{pages}{8193--8201}
  (\bibinfo{year}{2018}).

\bibitem{wang2020reinforcement}
\bibinfo{author}{Wang, Z.} \& \bibinfo{author}{Hong, T.}
\newblock \bibinfo{title}{Reinforcement learning for building controls: The
  opportunities and challenges}.
\newblock \emph{\bibinfo{journal}{Applied Energy}}
  \textbf{\bibinfo{volume}{269}}, \bibinfo{pages}{115036}
  (\bibinfo{year}{2020}).

\bibitem{bohacek2015molecular}
\bibinfo{author}{Bohacek, J.} \& \bibinfo{author}{Mansuy, I.~M.}
\newblock \bibinfo{title}{Molecular insights into transgenerational non-genetic
  inheritance of acquired behaviours}.
\newblock \emph{\bibinfo{journal}{Nat. Rev. Genet.}}
  \textbf{\bibinfo{volume}{16}}, \bibinfo{pages}{641--652}
  (\bibinfo{year}{2015}).

\bibitem{waddington1942canalization}
\bibinfo{author}{Waddington, C.~H.}
\newblock \bibinfo{title}{Canalization of development and the inheritance of
  acquired characters}.
\newblock \emph{\bibinfo{journal}{Nature}} \textbf{\bibinfo{volume}{150}},
  \bibinfo{pages}{563--565} (\bibinfo{year}{1942}).

\bibitem{andreev2022oldest}
\bibinfo{author}{Andreev, P.~S.} \emph{et~al.}
\newblock \bibinfo{title}{The oldest gnathostome teeth}.
\newblock \emph{\bibinfo{journal}{Nature}} \textbf{\bibinfo{volume}{609}},
  \bibinfo{pages}{964--968} (\bibinfo{year}{2022}).

\bibitem{zador2023catalyzing}
\bibinfo{author}{Zador, A.} \emph{et~al.}
\newblock \bibinfo{title}{Catalyzing next-generation artificial intelligence
  through neuroai}.
\newblock \emph{\bibinfo{journal}{Nat. Commun.}} \textbf{\bibinfo{volume}{14}},
  \bibinfo{pages}{1597} (\bibinfo{year}{2023}).

\bibitem{lillicrap2020backpropagation}
\bibinfo{author}{Lillicrap, T.~P.}, \bibinfo{author}{Santoro, A.},
  \bibinfo{author}{Marris, L.}, \bibinfo{author}{Akerman, C.~J.} \&
  \bibinfo{author}{Hinton, G.}
\newblock \bibinfo{title}{Backpropagation and the brain}.
\newblock \emph{\bibinfo{journal}{Nat. Rev. Neurosci.}}
  \textbf{\bibinfo{volume}{21}}, \bibinfo{pages}{335--346}
  (\bibinfo{year}{2020}).

\bibitem{zador2019critique}
\bibinfo{author}{Zador, A.~M.}
\newblock \bibinfo{title}{A critique of pure learning and what artificial
  neural networks can learn from animal brains}.
\newblock \emph{\bibinfo{journal}{Nat. Commun.}} \textbf{\bibinfo{volume}{10}},
  \bibinfo{pages}{3770} (\bibinfo{year}{2019}).

\bibitem{robinson2017epigenetics}
\bibinfo{author}{Robinson, G.~E.} \& \bibinfo{author}{Barron, A.~B.}
\newblock \bibinfo{title}{Epigenetics and the evolution of instincts}.
\newblock \emph{\bibinfo{journal}{Science}} \textbf{\bibinfo{volume}{356}},
  \bibinfo{pages}{26--27} (\bibinfo{year}{2017}).

\bibitem{moravec1988mind}
\bibinfo{author}{Moravec, H.}
\newblock \emph{\bibinfo{title}{Mind children: The future of robot and human
  intelligence}} (\bibinfo{publisher}{Harvard University Press},
  \bibinfo{year}{1988}).

\bibitem{poo2016china}
\bibinfo{author}{Poo, M.-m.} \emph{et~al.}
\newblock \bibinfo{title}{China brain project: basic neuroscience, brain
  diseases, and brain-inspired computing}.
\newblock \emph{\bibinfo{journal}{Neuron}} \textbf{\bibinfo{volume}{92}},
  \bibinfo{pages}{591--596} (\bibinfo{year}{2016}).

\bibitem{lv2022post}
\bibinfo{author}{Lv, Y.} \emph{et~al.}
\newblock \bibinfo{title}{Post-silicon nano-electronic device and its
  application in brain-inspired chips}.
\newblock \emph{\bibinfo{journal}{Front. Neurorobot.}}
  \textbf{\bibinfo{volume}{16}}, \bibinfo{pages}{1--17} (\bibinfo{year}{2022}).

\bibitem{mehonic2022brain}
\bibinfo{author}{Mehonic, A.} \& \bibinfo{author}{Kenyon, A.~J.}
\newblock \bibinfo{title}{Brain-inspired computing needs a master plan}.
\newblock \emph{\bibinfo{journal}{Nature}} \textbf{\bibinfo{volume}{604}},
  \bibinfo{pages}{255--260} (\bibinfo{year}{2022}).

\bibitem{gilbert2005genetic}
\bibinfo{author}{Gilbert, S.~L.}, \bibinfo{author}{Dobyns, W.~B.} \&
  \bibinfo{author}{Lahn, B.~T.}
\newblock \bibinfo{title}{Genetic links between brain development and brain
  evolution}.
\newblock \emph{\bibinfo{journal}{Nat. Rev. Genet.}}
  \textbf{\bibinfo{volume}{6}}, \bibinfo{pages}{581--590}
  (\bibinfo{year}{2005}).

\bibitem{roberts2022evolution}
\bibinfo{author}{Roberts, R.~J.}, \bibinfo{author}{Pop, S.} \&
  \bibinfo{author}{Prieto-Godino, L.~L.}
\newblock \bibinfo{title}{Evolution of central neural circuits: state of the
  art and perspectives}.
\newblock \emph{\bibinfo{journal}{Nat. Rev. Neurosci.}}
  \textbf{\bibinfo{volume}{23}}, \bibinfo{pages}{725--743}
  (\bibinfo{year}{2022}).

\bibitem{niven2016evolving}
\bibinfo{author}{Niven, J.~E.} \& \bibinfo{author}{Chittka, L.}
\newblock \bibinfo{title}{Evolving understanding of nervous system evolution}.
\newblock \emph{\bibinfo{journal}{Curr. Biol.}} \textbf{\bibinfo{volume}{26}},
  \bibinfo{pages}{R937--R941} (\bibinfo{year}{2016}).

\bibitem{brodner2018super}
\bibinfo{author}{Br{\"o}dner, P.}
\newblock \bibinfo{title}{{“Super-intelligent” machine: technological
  exuberance or the road to subjection}}.
\newblock \emph{\bibinfo{journal}{AI Soc}} \textbf{\bibinfo{volume}{33}},
  \bibinfo{pages}{335--346} (\bibinfo{year}{2018}).

\bibitem{luo2021architectures}
\bibinfo{author}{Luo, L.}
\newblock \bibinfo{title}{Architectures of neuronal circuits}.
\newblock \emph{\bibinfo{journal}{Science}} \textbf{\bibinfo{volume}{373}},
  \bibinfo{pages}{eabg7285} (\bibinfo{year}{2021}).

\bibitem{hasson2020direct}
\bibinfo{author}{Hasson, U.}, \bibinfo{author}{Nastase, S.~A.} \&
  \bibinfo{author}{Goldstein, A.}
\newblock \bibinfo{title}{Direct fit to nature: an evolutionary perspective on
  biological and artificial neural networks}.
\newblock \emph{\bibinfo{journal}{Neuron}} \textbf{\bibinfo{volume}{105}},
  \bibinfo{pages}{416--434} (\bibinfo{year}{2020}).

\bibitem{gupta2021embodied}
\bibinfo{author}{Gupta, A.}, \bibinfo{author}{Savarese, S.},
  \bibinfo{author}{Ganguli, S.} \& \bibinfo{author}{Fei-Fei, L.}
\newblock \bibinfo{title}{Embodied intelligence via learning and evolution}.
\newblock \emph{\bibinfo{journal}{Nat. Commun.}} \textbf{\bibinfo{volume}{12}},
  \bibinfo{pages}{5721} (\bibinfo{year}{2021}).

\bibitem{hoffmann2011climate}
\bibinfo{author}{Hoffmann, A.~A.} \& \bibinfo{author}{Sgr{\`o}, C.~M.}
\newblock \bibinfo{title}{Climate change and evolutionary adaptation}.
\newblock \emph{\bibinfo{journal}{Nature}} \textbf{\bibinfo{volume}{470}},
  \bibinfo{pages}{479--485} (\bibinfo{year}{2011}).

\bibitem{dickinson2000animals}
\bibinfo{author}{Dickinson, M.~H.} \emph{et~al.}
\newblock \bibinfo{title}{How animals move: an integrative view}.
\newblock \emph{\bibinfo{journal}{Science}} \textbf{\bibinfo{volume}{288}},
  \bibinfo{pages}{100--106} (\bibinfo{year}{2000}).

\bibitem{wang2022learngene}
\bibinfo{author}{Wang, Q.-F.} \emph{et~al.}
\newblock \bibinfo{title}{Learngene: From open-world to your learning task}.
\newblock In \emph{\bibinfo{booktitle}{Proceedings of the AAAI Conference on
  Artificial Intelligence}}, \bibinfo{pages}{8557--8565}
  (\bibinfo{year}{2022}).

\bibitem{schulman2017proximal}
\bibinfo{author}{Schulman, J.}, \bibinfo{author}{Wolski, F.},
  \bibinfo{author}{Dhariwal, P.}, \bibinfo{author}{Radford, A.} \&
  \bibinfo{author}{Klimov, O.}
\newblock \bibinfo{title}{Proximal policy optimization algorithms}.
\newblock \emph{\bibinfo{journal}{arXiv preprint arXiv:1707.06347}}
  (\bibinfo{year}{2017}).

\bibitem{jiang2020structure}
\bibinfo{author}{Jiang, X.}, \bibinfo{author}{Saggar, H.},
  \bibinfo{author}{Ryu, S.~I.}, \bibinfo{author}{Shenoy, K.~V.} \&
  \bibinfo{author}{Kao, J.~C.}
\newblock \bibinfo{title}{Structure in neural activity during observed and
  executed movements is shared at the neural population level, not in single
  neurons}.
\newblock \emph{\bibinfo{journal}{Cell Rep.}} \textbf{\bibinfo{volume}{32}},
  \bibinfo{pages}{108006} (\bibinfo{year}{2020}).

\bibitem{zhuang2020comprehensive}
\bibinfo{author}{Zhuang, F.} \emph{et~al.}
\newblock \bibinfo{title}{A comprehensive survey on transfer learning}.
\newblock \emph{\bibinfo{journal}{Proc. IEEE Inst. Electr. Electron. Eng.}}
  \textbf{\bibinfo{volume}{109}}, \bibinfo{pages}{43--76}
  (\bibinfo{year}{2020}).

\bibitem{pratt1992discriminability}
\bibinfo{author}{Pratt, L.~Y.}
\newblock \bibinfo{title}{Discriminability-based transfer between neural
  networks}.
\newblock \emph{\bibinfo{journal}{Proceedings of Advances in Neural Information
  Processing Systems}} \bibinfo{pages}{204--211} (\bibinfo{year}{1992}).

\bibitem{zoph2020rethinking}
\bibinfo{author}{Zoph, B.} \emph{et~al.}
\newblock \bibinfo{title}{Rethinking pre-training and self-training}.
\newblock In \emph{\bibinfo{booktitle}{Proceedings of Advances in Neural
  Information Processing Systems}}, \bibinfo{pages}{3833--3845}
  (\bibinfo{year}{2020}).

\bibitem{rosenstein2005transfer}
\bibinfo{author}{Rosenstein, M.~T.}, \bibinfo{author}{Marx, Z.},
  \bibinfo{author}{Kaelbling, L.~P.} \& \bibinfo{author}{Dietterich, T.~G.}
\newblock \bibinfo{title}{To transfer or not to transfer}.
\newblock In \emph{\bibinfo{booktitle}{NIPS 2005 Workshop on Transfer
  Learning}}, \bibinfo{pages}{1--4} (\bibinfo{year}{2005}).

\bibitem{wang2019characterizing}
\bibinfo{author}{Wang, Z.}, \bibinfo{author}{Dai, Z.},
  \bibinfo{author}{P{\'o}czos, B.} \& \bibinfo{author}{Carbonell, J.}
\newblock \bibinfo{title}{Characterizing and avoiding negative transfer}.
\newblock In \emph{\bibinfo{booktitle}{Proceedings of the IEEE Conference on
  Computer Vision and Pattern Recognition}}, \bibinfo{pages}{11293--11302}
  (\bibinfo{year}{2019}).

\bibitem{saxe2014exact}
\bibinfo{author}{Saxe, A.}, \bibinfo{author}{McClelland, J.} \&
  \bibinfo{author}{Ganguli, S.}
\newblock \bibinfo{title}{Exact solutions to the nonlinear dynamics of learning
  in deep linear neural networks}.
\newblock In \emph{\bibinfo{booktitle}{Proceedings of the International
  Conference on Learning Represenatations}}, \bibinfo{pages}{1--15}
  (\bibinfo{year}{2014}).

\bibitem{kimura1983neutral}
\bibinfo{author}{Kimura, M.}
\newblock \emph{\bibinfo{title}{The neutral theory of molecular evolution}}
  (\bibinfo{publisher}{Cambridge University Press}, \bibinfo{year}{1983}).

\bibitem{jablonka1998lamarckian}
\bibinfo{author}{Jablonka, E.}, \bibinfo{author}{Lamb, M.~J.} \&
  \bibinfo{author}{Avital, E.}
\newblock \bibinfo{title}{{`Lamarckian' mechanisms in Darwinian evolution}}.
\newblock \emph{\bibinfo{journal}{Trends Ecol. Evol.}}
  \textbf{\bibinfo{volume}{13}}, \bibinfo{pages}{206--210}
  (\bibinfo{year}{1998}).

\bibitem{larson2004evolution}
\bibinfo{author}{Larson, E.~J.} \emph{et~al.}
\newblock \emph{\bibinfo{title}{Evolution: The remarkable history of a
  scientific theory}}, vol.~\bibinfo{volume}{17} (\bibinfo{publisher}{Random
  House Digital, Inc.}, \bibinfo{year}{2004}).

\bibitem{gould2010lying}
\bibinfo{author}{Gould, S.~J.}
\newblock \emph{\bibinfo{title}{The lying stones of Marrakech: Penultimate
  reflections in natural history}} (\bibinfo{publisher}{Random House},
  \bibinfo{year}{2010}).

\bibitem{hughes2008ecological}
\bibinfo{author}{Hughes, A.~R.}, \bibinfo{author}{Inouye, B.~D.},
  \bibinfo{author}{Johnson, M.~T.}, \bibinfo{author}{Underwood, N.} \&
  \bibinfo{author}{Vellend, M.}
\newblock \bibinfo{title}{Ecological consequences of genetic diversity}.
\newblock \emph{\bibinfo{journal}{Ecol. Lett.}} \textbf{\bibinfo{volume}{11}},
  \bibinfo{pages}{609--623} (\bibinfo{year}{2008}).

\bibitem{watson1953molecular}
\bibinfo{author}{Watson, J.~D.} \& \bibinfo{author}{Crick, F.~H.}
\newblock \bibinfo{title}{Molecular structure of nucleic acids: a structure for
  deoxyribose nucleic acid}.
\newblock \emph{\bibinfo{journal}{Nature}} \textbf{\bibinfo{volume}{171}},
  \bibinfo{pages}{737--738} (\bibinfo{year}{1953}).

\bibitem{hochschild1986lambda}
\bibinfo{author}{Hochschild, A.}, \bibinfo{author}{Douhan~III, J.} \&
  \bibinfo{author}{Ptashne, M.}
\newblock \bibinfo{title}{How $\lambda$ repressor and $\lambda$ cro distinguish
  between or1 and or3}.
\newblock \emph{\bibinfo{journal}{Cell}} \textbf{\bibinfo{volume}{47}},
  \bibinfo{pages}{807--816} (\bibinfo{year}{1986}).

\bibitem{baker2008molecular}
\bibinfo{author}{Baker, T.~A.} \emph{et~al.}
\newblock \emph{\bibinfo{title}{Molecular biology of the gene}}
  (\bibinfo{publisher}{San Francisco, CA, USA:: Pearson/Benjamin Cummings},
  \bibinfo{year}{2008}).

\bibitem{todorov2012mujoco}
\bibinfo{author}{Todorov, E.}, \bibinfo{author}{Erez, T.} \&
  \bibinfo{author}{Tassa, Y.}
\newblock \bibinfo{title}{Mujoco: A physics engine for model-based control}.
\newblock In \emph{\bibinfo{booktitle}{Proceedings of IEEE/RSJ International
  Conference on Intelligent Robots and Systems}}, \bibinfo{pages}{5026--5033}
  (\bibinfo{year}{2012}).

\bibitem{konda1999actor}
\bibinfo{author}{Konda, V.} \& \bibinfo{author}{Tsitsiklis, J.}
\newblock \bibinfo{title}{Actor-critic algorithms}.
\newblock In \emph{\bibinfo{booktitle}{Proceedings of Advances in Neural
  Information Processing Systems}}, \bibinfo{pages}{1008--1014}
  (\bibinfo{year}{1999}).

\bibitem{brandt2013ancient}
\bibinfo{author}{Brandt, G.} \emph{et~al.}
\newblock \bibinfo{title}{Ancient dna reveals key stages in the formation of
  central european mitochondrial genetic diversity}.
\newblock \emph{\bibinfo{journal}{Science}} \textbf{\bibinfo{volume}{342}},
  \bibinfo{pages}{257--261} (\bibinfo{year}{2013}).

\bibitem{achtman2008microbial}
\bibinfo{author}{Achtman, M.} \& \bibinfo{author}{Wagner, M.}
\newblock \bibinfo{title}{Microbial diversity and the genetic nature of
  microbial species}.
\newblock \emph{\bibinfo{journal}{Nat. Rev. Microbiol.}}
  \textbf{\bibinfo{volume}{6}}, \bibinfo{pages}{431--440}
  (\bibinfo{year}{2008}).

\bibitem{guijarro2020widespread}
\bibinfo{author}{Guijarro-Clarke, C.}, \bibinfo{author}{Holland, P.~W.} \&
  \bibinfo{author}{Paps, J.}
\newblock \bibinfo{title}{Widespread patterns of gene loss in the evolution of
  the animal kingdom}.
\newblock \emph{\bibinfo{journal}{Nat. Ecol. Evol.}}
  \textbf{\bibinfo{volume}{4}}, \bibinfo{pages}{519--523}
  (\bibinfo{year}{2020}).

\bibitem{sharma2018genomics}
\bibinfo{author}{Sharma, V.} \emph{et~al.}
\newblock \bibinfo{title}{A genomics approach reveals insights into the
  importance of gene losses for mammalian adaptations}.
\newblock \emph{\bibinfo{journal}{Nat. Commun.}} \textbf{\bibinfo{volume}{9}},
  \bibinfo{pages}{1215} (\bibinfo{year}{2018}).

\bibitem{xue2023functional}
\bibinfo{author}{Xue, J.~R.} \emph{et~al.}
\newblock \bibinfo{title}{The functional and evolutionary impacts of
  human-specific deletions in conserved elements}.
\newblock \emph{\bibinfo{journal}{Science}} \textbf{\bibinfo{volume}{380}},
  \bibinfo{pages}{eabn2253} (\bibinfo{year}{2023}).

\end{thebibliography}

\begin{addendum}
 \item We sincerely thank Yongbiao Gao, Congzhi Zhang and Wenqian Li for the helpful discussion, and thank macrovector, brgfx, pikisuperstar and freepik for designing some figures. 
 \item[Competing Interests] The authors declare that they have no
competing financial interests.
 \item[Correspondence] Correspondence and requests for materials
should be addressed to Xin Geng (email: xgeng@seu.edu.cn).
\end{addendum}

\end{document}